\newcommand{\CLASSINPUTinnersidemargin}{10mm}
\documentclass[10pt,journal,compsoc]{IEEEtran}
\usepackage{booktabs}

\ifCLASSOPTIONcompsoc
  \usepackage[nocompress]{cite}
\else
  \usepackage{cite}
\fi

\usepackage{amsmath}
\usepackage{amssymb}
\usepackage{bm}
\usepackage{booktabs}
\usepackage{cases}
\usepackage{mdwmath}
\usepackage[dvipsnames]{xcolor}
\usepackage{array}
\usepackage{threeparttable}
\usepackage{color}
\usepackage{wasysym}
\usepackage{makecell}
\usepackage{dsfont}
\usepackage{bbm}
\usepackage{mathrsfs}
\usepackage{pifont}
\usepackage{marvosym}
\usepackage{colortbl}  %
\usepackage{xcolor}
\usepackage{multirow,diagbox}  %
\usepackage{float}
\usepackage{subfig}
\usepackage{ragged2e}
\usepackage{amsthm}
\usepackage{mathrsfs}
\usepackage{verbatim}

\newcommand{\myPara}[1]{\vspace{0.1in}\noindent\textbf{#1}}
\newcommand{\tabincell}[2]{\begin{tabular}{@{}#1@{}}#2\end{tabular}}
\newcommand{\cmark}{\ding{51}}%
\newcommand{\xmark}{\ding{55}}%

\def\ie{{\em i.e.}}
\def\eg{{\em e.g.}}

\makeatletter

\newcommand{\Rmnum}[1]{\expandafter\@slowromancap\romannumeral #1@}
\makeatother

\usepackage{tikz}
\usepackage{verbatim}
\usetikzlibrary{backgrounds,mindmap,trees}
\usetikzlibrary{calc,positioning,intersections}
\usepackage{pgfplots}
\usepackage{listings}
\usepackage{genealogytree}
\usepackage{mathrsfs}
\usepackage{multicol} 
\usepackage{hyperref}
\hypersetup{
    colorlinks=true,       
    linkcolor=blue,       
    filecolor=black,       
    urlcolor=black,        
    citecolor=blue,        
    pdftitle={Overleaf Example},
    pdfpagemode=FullScreen,
}

\usepackage{arydshln} 
\usepackage{threeparttable}
\usepackage{tabularx}
\usepackage{longtable}

\usepackage{utfsym}

\begin{document}

\title{How Far are Modern Trackers from UAV-Anti-UAV? A Million-Scale Benchmark and New Baseline}

\author{Chunhui~Zhang, Li~Liu,~\IEEEmembership{Senior Member,~IEEE,} Zhipeng~Zhang, Yong~Wang, Hao~Wen, Xi~Zhou, Shiming~Ge,~\IEEEmembership{Senior Member,~IEEE,} and Yanfeng~Wang 

\IEEEcompsocitemizethanks{
\IEEEcompsocthanksitem Chunhui~Zhang is with the Shanghai Jiao Tong University, Shanghai, 200240, China, the Hong Kong University of Science and Technology (Guangzhou), Guangzhou, 511458, China, and the CloudWalk Technology Co., Ltd, 201203, China. E-mail: chunhui.zhang@sjtu.edu.cn.\protect

\IEEEcompsocthanksitem Li~Liu is with the Hong Kong University of Science and Technology (Guangzhou), Guangzhou, 511458, China. E-mail: avrillliu@hkust-gz.edu.cn.\protect

\IEEEcompsocthanksitem Zhipeng Zhang and Yanfeng Wang are with the School of Artificial Intelligence, Shanghai Jiao Tong University, Shanghai, 200240, China. E-mails: zhipeng.zhang.cv@outlook.com, wangyanfeng622@sjtu.edu.cn.\protect

\IEEEcompsocthanksitem Yong~Wang is with the  School of Aeronautics and Astronautics, Sun Yat-sen University, Shenzhen, 518107, China. E-mail: wangyong5@mail.sysu.edu.cn.\protect

\IEEEcompsocthanksitem Hao~Wen and Xi Zhou are with the CloudWalk Technology Co., Ltd, 201203, China. E-mails: \{wenhao, zhouxi\}@cloudwalk.com.\protect

\IEEEcompsocthanksitem Shiming Ge is with the Institute of Information Engineering, Chinese Academy of Sciences, Beijing 100085, China. E-mail: geshiming@iie.ac.cn.\protect
\vspace{0.3cm}


}
\thanks{This work was done while interning at the Hong Kong University of Science and Technology (Guangzhou).}
\thanks{{\normalsize \Letter}~Corresponding author: Li Liu.}
\thanks{\color{blue}This technical report presents an initial evaluation of 50 representative modern trackers and details our methodology and experimental setup. Results will be updated in future versions as the implementation evolves.}
}

\markboth{IEEE Transactions XXX,~Vol.~XX, No.~XX, December~2025}%
{Shell \MakeLowercase{\textit{et al.}}: Bare Advanced Demo of IEEEtran.cls for IEEE Computer Society Journals}

\IEEEtitleabstractindextext{%
\begin{abstract}
Unmanned Aerial Vehicles (UAVs) offer wide-ranging applications but also pose significant safety and privacy violation risks in areas like airport and infrastructure inspection, spurring the rapid development of Anti-UAV technologies in recent years. However, current Anti-UAV research primarily focuses on RGB, infrared (IR), or RGB-IR videos captured by fixed ground cameras, with little attention to tracking target UAVs from another moving UAV platform. To fill this gap, we propose a new multi-modal visual tracking task termed \emph{UAV-Anti-UAV}, which involves a pursuer UAV tracking a target adversarial UAV in the video stream. Compared to existing Anti-UAV tasks, UAV-Anti-UAV is more challenging due to severe dual-dynamic disturbances caused by the rapid motion of both the capturing platform and the target. To advance research in this domain, we construct a million-scale dataset consisting of 1,810 videos, each manually annotated with bounding boxes, a language prompt, and 15 tracking attributes. Furthermore, we propose MambaSTS, a Mamba-based baseline method for UAV-Anti-UAV tracking, which enables integrated spatial-temporal-semantic learning. Specifically, we employ Mamba and Transformer models to learn global semantic and spatial features, respectively, and leverage the state space model’s strength in long-sequence modeling to establish video-level long-term context via a temporal token propagation mechanism. We conduct experiments on the UAV-Anti-UAV dataset to validate the effectiveness of our method. A thorough experimental evaluation of 50 modern deep tracking algorithms demonstrates that there is still significant room for improvement in the UAV-Anti-UAV domain. The dataset and codes will be available at {\color{magenta}https://github.com/983632847/Awesome-Multimodal-Object-Tracking}.

\end{abstract}

\begin{IEEEkeywords}
UAV-Anti-UAV, Benchmark, Mamba Model, Vision-Language Tracking, Modern Trackers, Low-Altitude Security
\end{IEEEkeywords}}

\maketitle
\IEEEdisplaynontitleabstractindextext
\IEEEpeerreviewmaketitle

\section{Introduction}
\label{sec:introduction}

Unmanned Aerial Vehicles (UAVs) have rapidly advanced in recent years and are widely deployed in diverse industrial and civil applications, including environmental monitoring, logistics, transportation, agriculture, and public safety surveillance~\cite{zhu2020vision,zhang2022webuav,mueller2016benchmark,du2018unmanned,zhang2024awesome}. However, this rapid adoption simultaneously introduces critical challenges to low-altitude security, particularly in safeguarding sensitive airspaces such as airports, power plants, and government facilities from unauthorized or malicious UAV incursions. As a consequence, Anti-UAV technologies have received significant research attention in both academia and industry, aiming to detect, track, and intercept adversarial UAVs~\cite{jiang2021anti,huang2023anti,xu2025tri,zhao2022vision}. Among these components, visual tracking plays a fundamental role because it provides continuous spatio-temporal state estimation essential for trajectory reasoning and interception strategies~\cite{javed2022visual,wu2015otb,smeulders2013visual,tang2025omni}.

\begin{figure}[t]
  \centering
\includegraphics[width=1.0\linewidth]{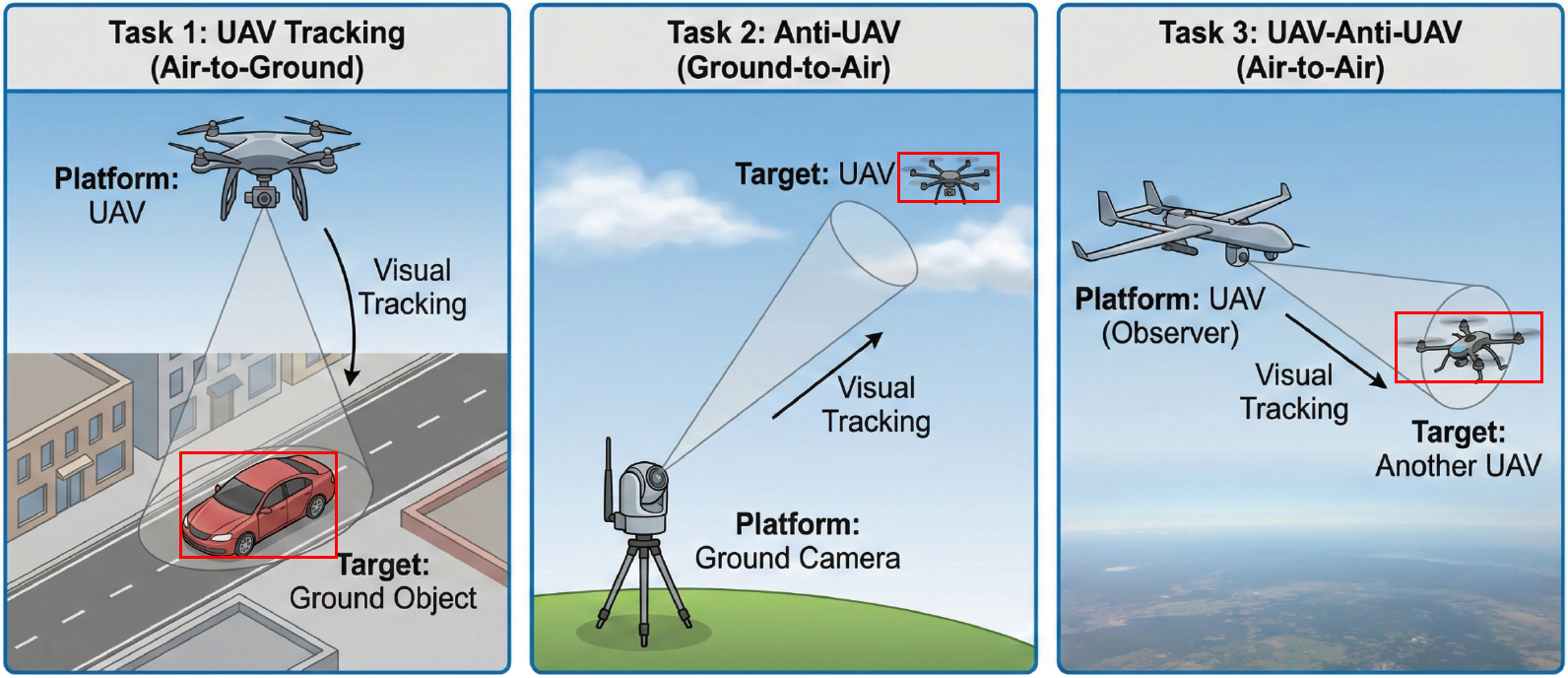}
  \caption{Comparison of three distinct UAV-related visual tracking tasks. (a) UAV Tracking: A UAV tracks ground targets (\eg, cars, pedestrians), characterized by top-down views and scale variations. (b) Anti-UAV: A ground-based camera tracks an airborne UAV, often facing cluttered sky backgrounds and tiny targets. (c) Proposed UAV-Anti-UAV: A chasing UAV tracks a target UAV. This task involves highly dynamic relative motion and erratic background changes due to the rapid movement of both the platform and the target.}
  \label{fig:motivation}
\end{figure}

\begin{figure*}[t]
  \centering
\includegraphics[width=1.0\linewidth]{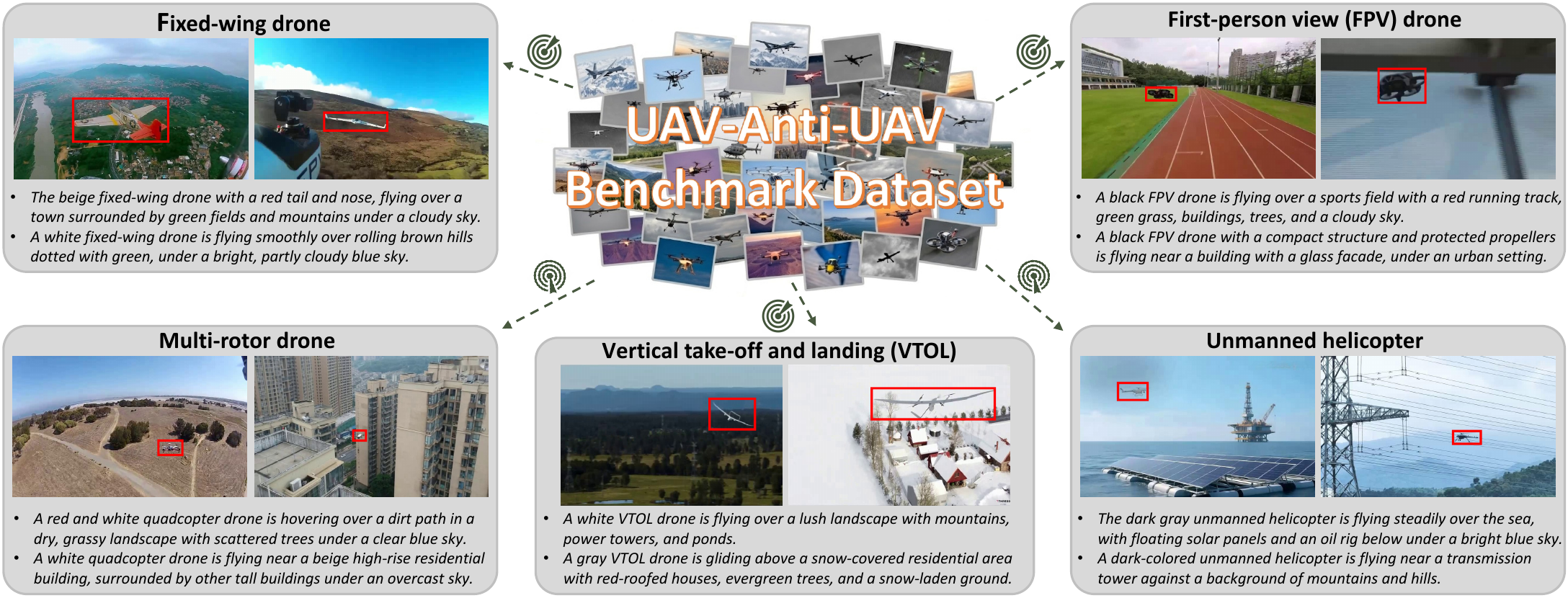}
  \caption{Representative examples from the UAV-Anti-UAV benchmark dataset. The dataset contains five distinct categories of target UAVs: fixed-wing, first-person view (FPV), multi-rotor, vertical take-off and landing (VTOL), and unmanned helicopter. Each example is annotated with bounding boxes and a corresponding language prompt describing the target and its environment.}
  \label{fig:examples}
\end{figure*}

Existing research on aerial visual tracking has been largely driven by publicly available UAV tracking benchmarks such as UAV123~\cite{mueller2016benchmark}, UAV20L~\cite{mueller2016benchmark}, DTB70~\cite{li2017visual}, UAVDT~\cite{du2018unmanned}, VisDrone~\cite{zhu2020vision}, UAVDark135~\cite{li2021all}, and WebUAV-3M~\cite{zhang2022webuav}. These datasets investigate typical tracking challenges—such as viewpoint variation, scale changes, occlusion, illumination fluctuation, and motion blur—from the perspective of a UAV tracking ground targets. However, as shown in Fig.~\ref{fig:motivation}(a), these datasets model a UAV-to-ground tracking paradigm, where the capturing platform moves while the target object remains mostly motion-stable relative to the ground. As a result, they do not fully reflect real-world aerial confrontation environments.

To address threats from hostile UAVs, recent Anti-UAV datasets such as Anti-UAV318~\cite{jiang2021anti}, DUT Anti-UAV~\cite{zhao2022vision}, Anti-UAV600~\cite{zhu2023evidential}, Anti-UAV410~\cite{huang2023anti}, and MM-UAV~\cite{xu2025tri} adopt a ground-to-UAV tracking paradigm, in which a static ground-based camera observes flying UAVs from a fixed or limited-motion viewpoint (see Fig.~\ref{fig:motivation}(b)). These datasets improve the applicability of tracking models to airborne targets, yet they still suffer from limited motion diversity. The camera remains static or nearly static, meaning there is an absence of dual dynamic disturbances, platform vibration, aggressive viewpoint transitions, and high-speed close-range pursuit that characterize real aerial confrontation.

To bridge this gap between practical scenarios and existing research assumptions, we propose a new UAV-Anti-UAV tracking task, illustrated in Fig.~\ref{fig:motivation}(c). In this configuration, a pursuer UAV attempts to continuously track an adversarial UAV during a high-speed, low-altitude flight. This scenario introduces significantly greater complexity and realism than previous UAV tracking or Anti-UAV tracking. Specifically, both the tracker and target are in rapid motion, leading to: (1) severe dual dynamic disturbances, (2) rapid scale variation and very small-object ratios, (3) strong motion blur and turbulence, (4) frequent viewpoint switching due to platform rotation, and (5) rapid background changes. These conditions severely degrade the performance of modern trackers, including correlation-filter-based approaches~\cite{danelljan2017eco,mayer2022transforming,zhang2020accurate,danelljan2019atom,ge2020cascaded,zhang2019robust,ge2019distilling}, deep CNN trackers~\cite{li2019siamrpn++,ChenZLZJ20,guo2020siamcar,guo2021graph,bertinetto2016fully,XuWLYY20,song2018vital,wang2019fast,yan2021lighttrack}, Transformer-based models~\cite{ye2022joint,chen2022backbone,cui2023mixformerv2,li2023adaptive,gao2023generalized,gao2022aiatrack,cao2021hift,yan2021learning,chen2021transformer,wang2021transformer,lin2024tracking,wu2025learning,bai2024artrackv2,chen2023seqtrack}, Mamba-based sequence models~\cite{wu2024mambanut,li2025mambalct,kang2025exploring,zhang2025mambatrack}, and recently emerging  multi-modal vision-language trackers~\cite{guo2022divert,zhou2023joint,li2023citetracker,zhang2023all,ma2024unifying,chen2025sutrack,li2025dynamic,feng2025atctrack}. Despite their advancements, these methods were primarily evaluated under limited dynamic environments, making them unsuitable for real aerial adversarial pursuit~\cite{xu2025tri,dong2025securing}.

To advance research in this critical domain, we introduce the first million-scale UAV-Anti-UAV benchmark, consisting of 1,810 flight confrontation videos and around 1.05M densely annotated bounding boxes, including natural language prompts and 15 attribute categories representing realistic disturbance conditions. Additionally, we propose MambaSTS, a novel baseline that leverages \textbf{Mamba} state-space modeling~\cite{gu2023mamba} to achieve integrated \textbf{S}patial-\textbf{T}emporal-\textbf{S}emantic learning. Distinct from standard bidirectional state space models~\cite{zhu2024vision}, MambaSTS adopts a unidirectional scanning mechanism tailored for the causal nature of video tracking. This design facilitates a temporal token propagation mechanism, which efficiently compresses historical target cues—ranging from trajectory evolution to semantic descriptions—into a compact latent state. By integrating these video-level long-term contexts with spatial and language features, MambaSTS effectively mitigates the severe dual-dynamic disturbances inherent in UAV-Anti-UAV tracking. We further conduct a comprehensive evaluation of 50 state-of-the-art (SOTA) deep trackers from different paradigms, revealing a substantial performance gap between current trackers and the requirements of real confrontation scenarios.

\begin{table*}[ht]
{
\renewcommand\arraystretch{1.0}
\caption{Comparison of the proposed UAV-Anti-UAV dataset with popular UAV tracking and Anti-UAV tracking datasets.}
	\label{tab:Comp_UAVAntiUAV}
	\begin{center}
		\setlength{\tabcolsep}{1.21mm}{
			\scalebox{1.0}{
			\begin{tabular}{lcccccccccccccc}
				\Xhline{0.75pt} 
				Dataset & Year &  Videos  & Attributes  &   \tabincell{c}{ Min \\ frame} & \tabincell{c}{Mean\\ frame} & \tabincell{c}{Max\\frame} & \tabincell{c}{Total\\ frames}  &  \tabincell{c}{Total \\duration} & \tabincell{c}{Data \\partition}  &  \tabincell{c}{Absent\\ label} &\tabincell{c}{ Language \\prompt} 
                 & \tabincell{c}{Including \\UAV}   & \tabincell{c}{Capture \\by UAV} \\
				
				\hline
    
	        	\textbf{UAV123}~\cite{mueller2016benchmark} &  2016 & 123  & 12 & 109  & 915  & 3,085 & 113 K  & 62.5 min & Test & \xmark   & \xmark & \ding{52}\rotatebox[origin=c]{-9.2}{\kern-0.7em\ding{55}}  & \cmark \\	        	
                \textbf{UAV20L}~\cite{mueller2016benchmark} &  2016 & 20  & 12 & 1,717 & 2,934 & 5,527 & 59 K  & 32.6 min  & Test & \xmark   & \xmark & \ding{52}\rotatebox[origin=c]{-9.2}{\kern-0.7em\ding{55}} & \cmark \\	

                \textbf{DTB70}~\cite{li2017visual} &  2017 & 70  & 11 & 68 & 225 & 699 & 15.8 K   & 8.8 min & Test & \xmark   & \xmark & \xmark & \cmark \\	
                
                \textbf{UAVDT}~\cite{du2018unmanned} &  2018 & 50  &  8  & 82 & 742 & 2,969  & 37.1 K    & 20.6 min & Test & \xmark   & \xmark & \xmark & \cmark \\

                \textbf{VisDrone}~\cite{zhu2020vision} &  2021 & 167  & 12 & 90 & 834 & 4,280 & 139.28 K    & 77.4 min & Train/Val/Test & \xmark   & \xmark & \xmark & \cmark \\
                
                \textbf{UAVDark135}~\cite{li2021all} &  2022 & 135  & 12 & 216 & 929 & 4,571  &  125.47 K    & 69.7 min  & Test & \xmark   & \xmark & \xmark & \cmark \\

                \textbf{WebUAV-3M}~\cite{zhang2022webuav} &  2023 & 4,500  & 17 & 40 & 710 & 18,841 & 3.30 M  & 28.9 hours & Train/Val/Test & \cmark  & \cmark  & \ding{52}\rotatebox[origin=c]{-9.2}{\kern-0.7em\ding{55}}  & \cmark \\	
                
				\hline
                
	        	\textbf{Anti-UAV318}~\cite{jiang2021anti} &  2021 &  318  & 7 & 28 & 994 & 1,000 & 296.9 K  & 4.1 hours & Train/Val/Test & \cmark   & \xmark & \cmark & \xmark  \\	
 
                \textbf{DUT Anti-UAV}~\cite{zhao2022vision} &  2022 & 20  & - & 83 & 1,240 & 2,635  &  24.8 K    &  13.8 min  & Test & \xmark   & \xmark & \cmark & \xmark  \\	

                \textbf{Anti-UAV600}~\cite{zhu2023evidential} &  2023 & 600  & 7 & 28 & 1,158 & 1,500 & 723 K    & 6.7 hours  & Train/Val & \xmark   & \xmark & \cmark & \xmark  \\	
                
                \textbf{Anti-UAV410}~\cite{huang2023anti} &  2024  & 410  & 10 & 28 & 1,069 & 1,500 & 438 K    &  4.1 hours  & Train/Val/Test & \xmark   & \xmark & \cmark & \xmark  \\	
                
                \textbf{MM-UAV}~\cite{xu2025tri} &  2025 & 1,321  & 7 & 1,509 & 2,129 & 3,558 & 2.8 M   & 25.9 hours & Train/Test & \xmark & \xmark & \cmark & \xmark \\	

				\hline
				\textbf{UAV-Anti-UAV}~ & 2025 &  1,810  & 15  & 14 & 578 &  17,740 & 1.05 M  & 9.85 hours  & Train/Test & \cmark  & \cmark  & \cmark  & \cmark  \\
				\Xhline{0.75pt} 
			\end{tabular}
	   }
		}
	\end{center}
    }
        \begin{tablenotes}
        \footnotesize
        \item[] ``\ding{52}\rotatebox[origin=c]{-9.2}{\kern-0.7em\ding{55}}'' denotes that the corresponding dataset contains a small number of UAV videos.
    \end{tablenotes}
\end{table*}

In summary, our main contributions are fourfold:

\begin{itemize}
\item We introduce the UAV-Anti-UAV tracking task, establishing a new research direction for air-to-air dynamic visual tracking and highlighting its unique challenges compared to existing paradigms.

\item We construct the first large-scale UAV-Anti-UAV dataset, comprising 1,810 video sequences (1.05 million frames) captured in diverse real-world scenarios. Each sequence includes precise bounding box annotations, language prompts describing the target and scene, and 15 fine-grained tracking attributes (\eg, motion blur, small target, occlusion) to enable comprehensive evaluation.

\item We propose MambaSTS, a Mamba-Transformer hybrid framework that integrates spatial, temporal, and semantic learning. A lightweight STS Mamba module models long-term temporal dependencies via a temporal token propagation mechanism, while Transformer and Mamba backbones capture global spatial and semantic features, enabling robust tracking under dual-dynamic disturbances.

\item We evaluate 50 SOTA deep trackers across five categories (CNN-based, CNN-Transformer-based, Transformer-based, Mamba-based, and Transformer-Mamba-based) on our dataset. Results reveal significant performance gaps, particularly in handling challenging attributes like illumination variations and similar distractors, providing actionable insights for future research.
\end{itemize}

\section{Related Work}
\label{sec:related_work}

\subsection{Modern Deep Tracking} 

Visual tracking has evolved rapidly with deep learning, moving from handcrafted features to data-driven representations. Below, we briefly review four types of unimodal trackers and recently emerging vision-language tracking models. Readers can refer to~\cite{javed2022visual,xu2025tri,zhang2024awesome} to track the latest developments in visual tracking.

\myPara{CNN-based Trackers.} Early deep trackers leveraged convolutional neural networks (CNNs) for discriminative feature extraction. SiamFC~\cite{bertinetto2016fully} introduced the Siamese network architecture, learning similarity metrics between template and search regions to achieve real-time performance. Subsequent works like SiamRPN++~\cite{li2019siamrpn++}, AutoMatch~\cite{zhang2021learn}, SiamBAN~\cite{ChenZLZJ20}, SiamCAR~\cite{guo2020siamcar}, and SiamFC++~\cite{XuWLYY20} enhanced accuracy by integrating deeper backbones (\eg, ResNet-50~\cite{he2016deep}) and multi-scale feature fusion. ATOM~\cite{danelljan2019atom} and ECO~\cite{danelljan2017eco} optimized for precision and efficiency, respectively. However, these methods are limited by CNNs’ local receptive fields, failing to capture global context or long-term temporal dependencies—critical for dynamic aerial scenarios~\cite{zhang2020accurate,cao2023towards}.

\myPara{CNN-Transformer and Transformer-based Trackers.} The Transformer’s self-attention mechanism revolutionized tracking by modeling global spatial relationships~\cite{dosovitskiy2020image}. TrDiMP~\cite{wang2021transformer}, TransT~\cite{chen2021transformer}, and STARK~\cite{yan2021learning} integrated the Transformer with CNNs (\eg, ResNet-50), while Aba-ViTrack~\cite{li2023adaptive} proposed an adaptive and background-aware vision Transformer for real-time UAV tracking. OSTrack~\cite{ye2022joint}, SimTrack~\cite{chen2022backbone}, and MixFormerV2~\cite{cui2023mixformerv2} further improved feature learning via joint relation modeling and multi-scale aggregation. Despite their success on conventional benchmarks (\eg, LaSOT~\cite{fan2019lasot}), these trackers struggle with high computational complexity and inefficient temporal modeling, making them unsuitable for UAV-Anti-UAV’s dual-dynamic motion.

\myPara{Mamba-based Trackers.} Mamba~\cite{gu2023mamba}, a state space model (SSM) designed for long-sequence processing, has recently emerged as a promising alternative to Transformers. MambaTrack~\cite{zhang2025mambatrack}, MambaNUT~\cite{wu2024mambanut}, MambaLCT~\cite{li2025mambalct}, and MCITrack~\cite{kang2025exploring} applied Mamba to nighttime UAV tracking and general object tracking, demonstrating efficient temporal dependency modeling. However, existing Mamba-based trackers~\cite{li2025mambalct,kang2025exploring} mainly focus on temporal dynamics and lack integration of spatial-temporal-semantic information—essential for handling target appearance variations and similar distractors in aerial scenes.

\myPara{Vision-Language Trackers.} Vision-language (VL) trackers integrate natural language prompts to enhance semantic understanding, improving robustness against appearance changes~\cite{zhou2023joint,li2025dynamic,feng2025atctrack,wang2021towards,fan2019lasot}. VLT$_{\rm TT}$~\cite{guo2022divert} and CiteTracker~\cite{li2023citetracker} align visual features with pre-trained language models (\eg, CLIP~\cite{radford2021learning}, BERT~\cite{devlin2018bert}) to capture target semantics. All-in-One~\cite{zhang2023all} and UVLTrack~\cite{ma2024unifying} proposed multi-modal alignment for unified tracking, but these methods have not been evaluated in air-to-air dynamic scenarios, where semantic information must be combined with temporal motion modeling.

\subsection{UAV Tracking}

UAV tracking focuses on ground target localization using UAV-mounted cameras, addressing unique challenges like top-down viewpoints, scale variations, and motion blur~\cite{zhang2022webuav,mueller2016benchmark}. Some key datasets and methods include:

\myPara{Datasets:} UAV123~\cite{mueller2016benchmark}, UAV20L~\cite{mueller2016benchmark}, and DTB70~\cite{li2017visual} provided initial datasets focusing on generic objects. UAVDT~\cite{du2018unmanned} introduced more challenging urban traffic scenarios. VisDrone~\cite{zhu2020vision} significantly expanded the scale and diversity. WebUAV-3M~\cite{zhang2022webuav} further pushed the boundary with millions of frames. Specialized datasets like UAVDark135~\cite{li2021all} address the critical challenge of nighttime UAV tracking. These benchmarks have driven the development of numerous UAV-specific trackers tackling issues like extreme camera motion and small targets.

\myPara{Methods:} TCTrack~\cite{cao2022tctrack} and TCTrack++~\cite{cao2023towards} optimized temporal context modeling for aerial tracking, while HiFT~\cite{cao2021hift} used hierarchical feature Transformers to handle scale variations. Nighttime UAV tracking has also gained attention, with MambaNUT~\cite{wu2024mambanut} and MambaTrack~\cite{zhang2025mambatrack} proposing adaptive curriculum learning and dual-enhancement to improve low-visibility performance. However, these methods are designed for ground targets and lack proven success in handling the dual-dynamic motion characteristic of UAV-Anti-UAV scenarios.

\begin{table}[t]
{
   \caption{\footnotesize Definition of 15 attributes in the UAV-Anti-UAV dataset.}
    \label{tab:Attribute_defination}
	\begin{center}
		\setlength{\tabcolsep}{0.2mm}{
			\begin{tabular}{l|l}
				\Xhline{0.75pt} 
                
				\textbf{Attribute} & \textbf{Definition} \\
				\hline
                \textbf{01. CM} &  Abrupt motion of the camera. \\
                    
                \textbf{02. VC} &  Viewpoint affects target appearance significantly. \\

			    \textbf{03. PO} & The target is partially occluded in the sequence. \\
			    
                \textbf{04. FO} &  The target is fully occluded in the sequence. \\
                
                \textbf{05. OV} &  The target completely leaves the video frame. \\

                \textbf{06. ROT} & The target rotates in the video sequence. \\
                
                \textbf{07. SD} & There is a similar object or background near the target object.\\

                \textbf{08. IV} &  The illumination in the target region changes. \\
                
                \textbf{09. MB} & \tabincell{l}{The target region is blurred due to the target or camera motion.} \\

                \textbf{10. PTI} & Only part of the target information is visible in the initial frame.\\ 

                \textbf{11. SO} & \tabincell{l}{The target with an average relative size smaller than 1\% of image \\and an average absolute size less than $\sqrt{22 \times 22}$ pixels.}\\ 
                 
			    \textbf{12. FM} &  \tabincell{l}{The motion of the object is larger than its size.} \\
			    
                \textbf{13. SV} &  The ratio of the bounding box is outside the range [0.5, 2]. \\
                
                \textbf{14. ARV} &  \tabincell{l}{The ratio of bounding box aspect ratio is outside the range [0.5, 2].} \\
                
                \textbf{15. LEN} & \tabincell{l}{The length ($l$) of current video is short ($l \!\leq \!600$ frames, 20s for \\30 fps), or medium ($600 \!<\! l \!\leq\! 1800$ frames, 60s for 30 fps), \\or long ($l \!>\!1800$ frames).}\\
                
			\Xhline{0.75pt} 
			\end{tabular}
		}
	\end{center}
    }
\end{table}

\begin{figure}[t]   
	\centering\centerline{\includegraphics[width=1.0\linewidth]{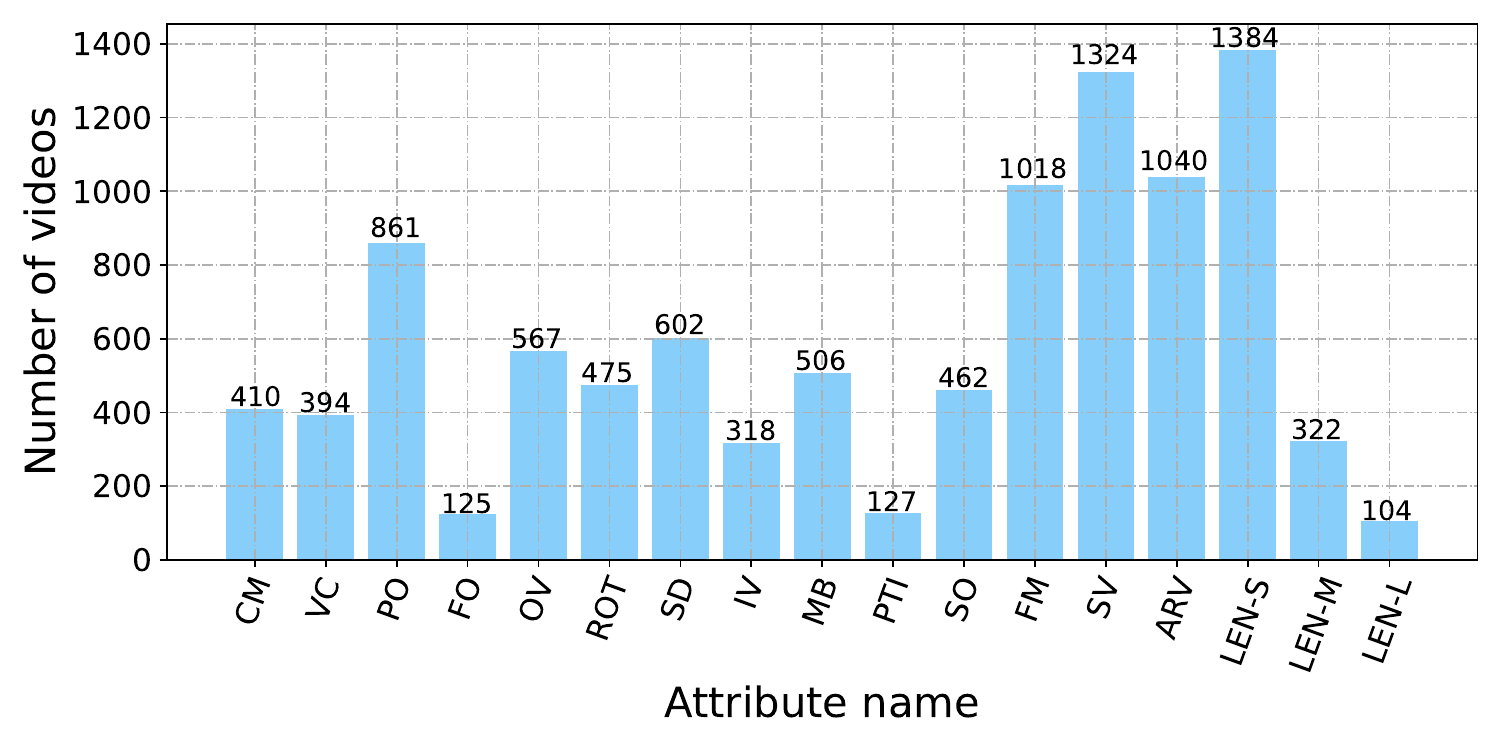}}
	\caption{Distribution of videos for each tracking attribute.}
	\label{fig:attribute_of_videos}
\end{figure}

\begin{figure}[t]   
\centering\centerline{\includegraphics[width=1.0\linewidth]{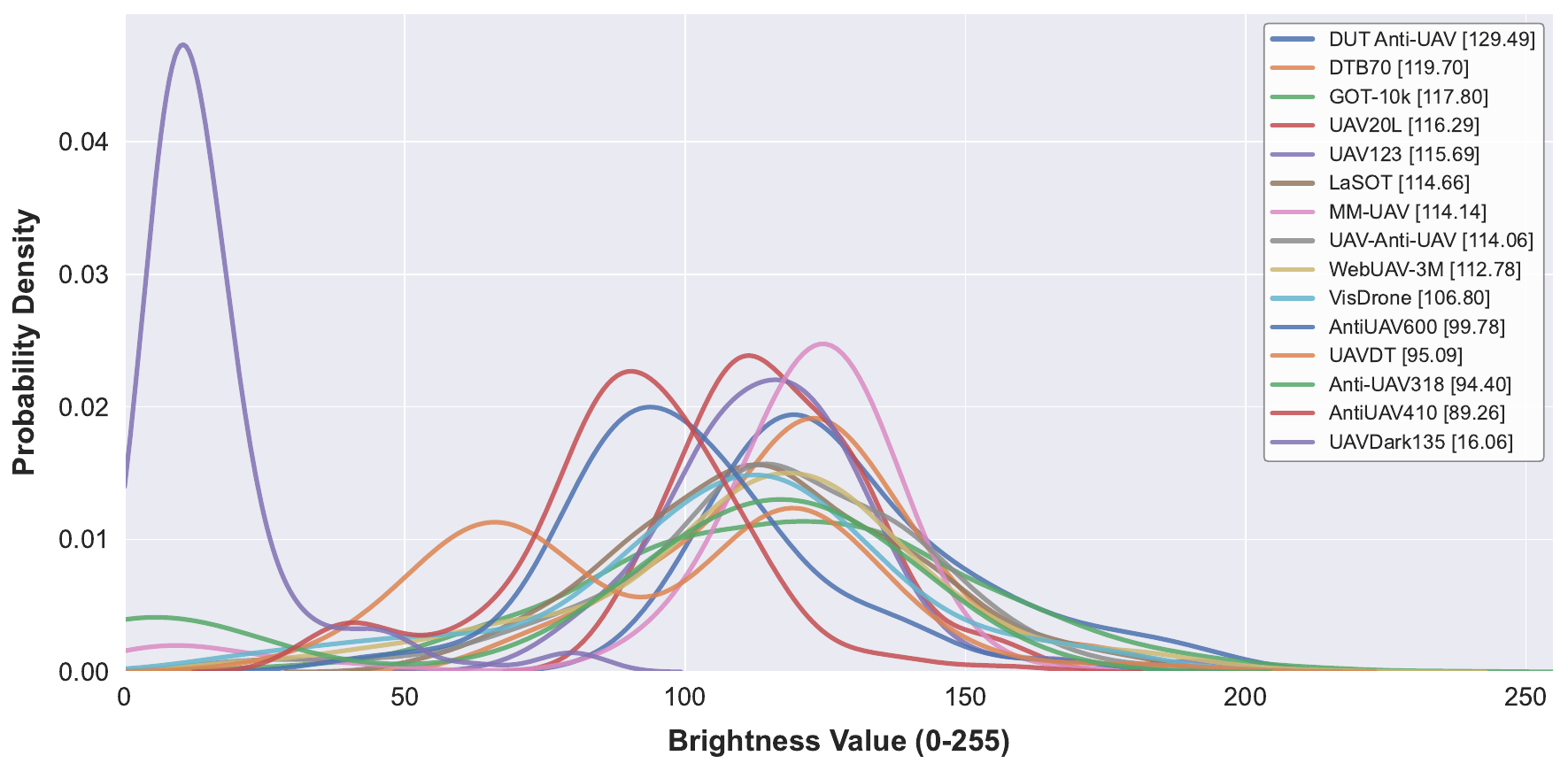}}
	\caption{Distribution of brightness values for the proposed dataset versus existing benchmarks. The average brightness of each dataset is provided in the legend.}
	\label{fig:brightness_comparison}
\end{figure}

\begin{figure}[t]   
\centering\centerline{\includegraphics[width=1.0\linewidth]{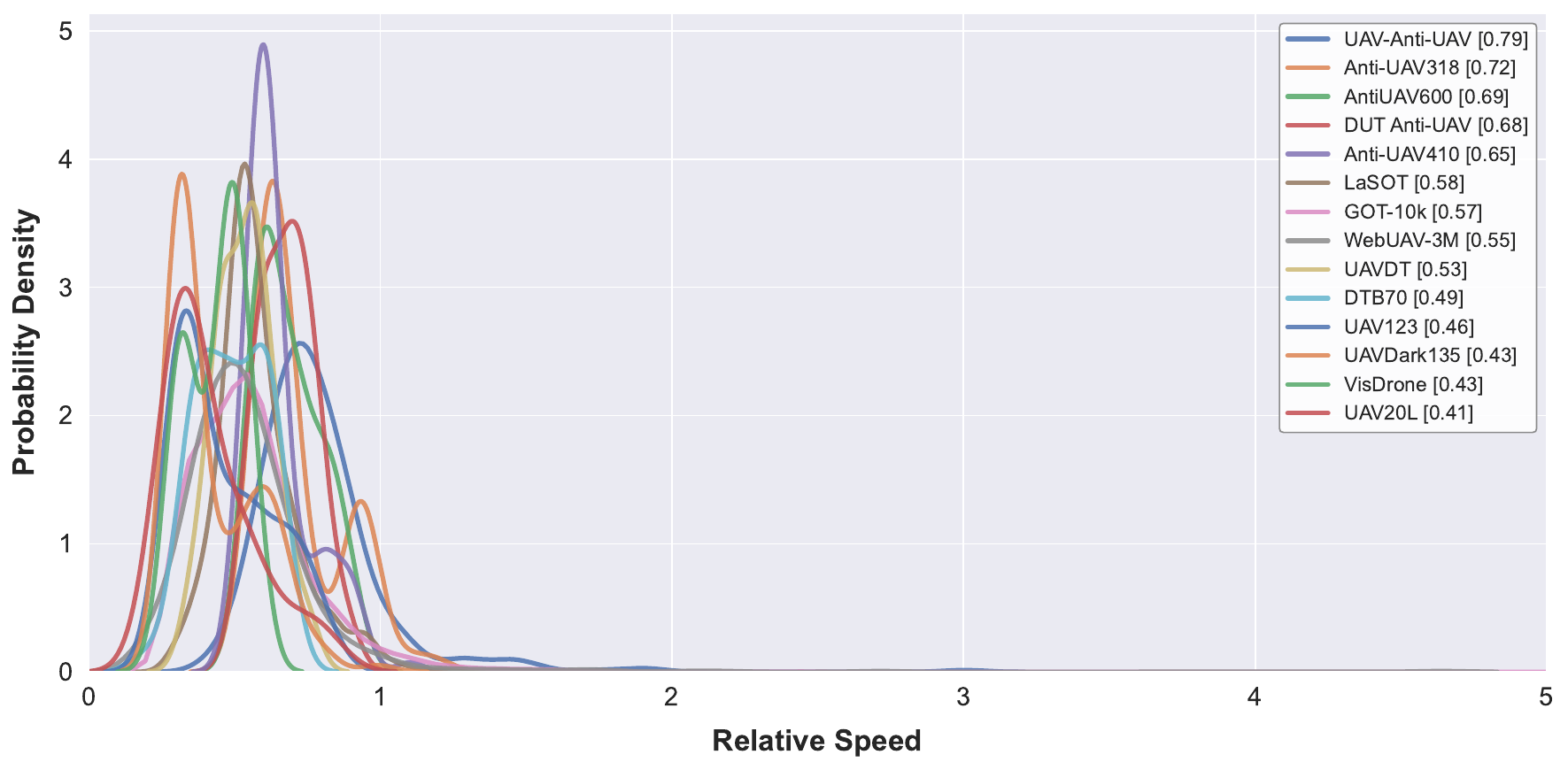}}
	\caption{Comparison of relative speed distributions between the proposed UAV-Anti-UAV dataset and existing UAV tracking and Anti-UAV datasets. The legend displays the average relative speed of each dataset. }
	\label{fig:average_relative_speed_distribution}
\end{figure}

\subsection{Anti-UAV Tracking}

Anti-UAV tracking is concerned with detecting and tracking UAVs from ground-based platforms, a core component of low-altitude security systems~\cite{jiang2021anti,zhao2022vision,zhu2023evidential,huang2023anti}.

\myPara{Datasets and Methods:} The Anti-UAV318 dataset~\cite{jiang2021anti} was a pioneering effort, providing sequences with small UAVs against complex skies. Follow-up datasets like DUT Anti-UAV~\cite{zhao2022vision}, Anti-UAV600~\cite{zhu2023evidential}, and Anti-UAV410~\cite{huang2023anti} increased the volume and diversity. The latest MM-UAV dataset~\cite{xu2025tri} incorporated multi-modal (RGB, infrared, and event) data. To address the challenges of tiny scales and rapid motion in these datasets, methodologies have evolved along three primary trajectories: (1) Adaptive Search Strategies, exemplified by FocusTrack~\cite{wang2025focustrack}, which dynamically adjusts the field-of-view to prevent target loss, and a detector-based approach~\cite{peng2025simple} that integrates frame dynamics and trajectory-constrained filtering to capture motion cues. Furthermore, SiamDT~\cite{huang2023anti} specifically tackles tiny thermal targets by introducing a dual-semantic feature extraction mechanism—combining matching semantics with foreground probability—and a background suppression branch to filter dynamic clutter. (2) Semantic and Language Integration, where DFSC~\cite{jiang2021anti} exploits class-level consistency, UAUTrack~\cite{ren2025uautrack} utilizes text prior prompts for unified multi-modal tracking, and JTD-UAV~\cite{wang2025jtd} leverages multimodal large language models for joint tracking and behavior description; and (3) Multi-modal Fusion, such as MMA-SORT~\cite{xu2025tri}, which employs offset-guided adaptive alignment to synergize RGB, infrared, and event streams. Despite these advancements, these methods predominantly assume a static or quasi-static observer, leaving the challenges of dual-dynamic air-to-air tracking largely unexplored.

\myPara{The Gap:} A fundamental limitation of all existing Anti-UAV research is the assumption of a static or ground-based observer. As Fig.~\ref{fig:motivation} illustrates, this differs drastically from the UAV-Anti-UAV scenario where both observer and target are dynamic aerial platforms. This shift introduces a new regime of challenges not addressed by current datasets or methods, creating the gap our work aims to fill.

 \begin{figure*}[t]
\vspace{-0.2cm}
\begin{minipage}[t]{0.542\linewidth}
\centering
\includegraphics[width=0.32\linewidth]{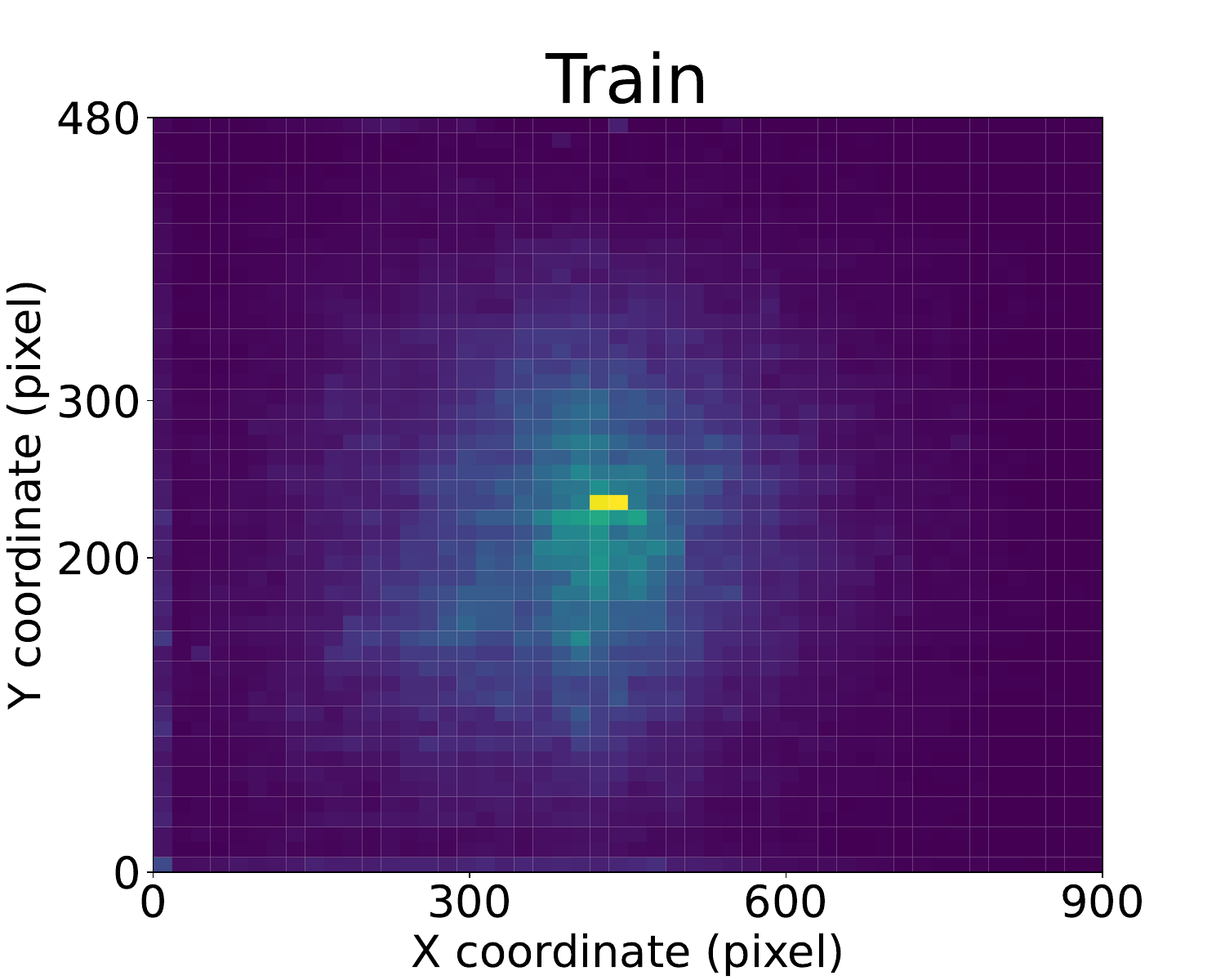}
\includegraphics[width=0.32\linewidth]{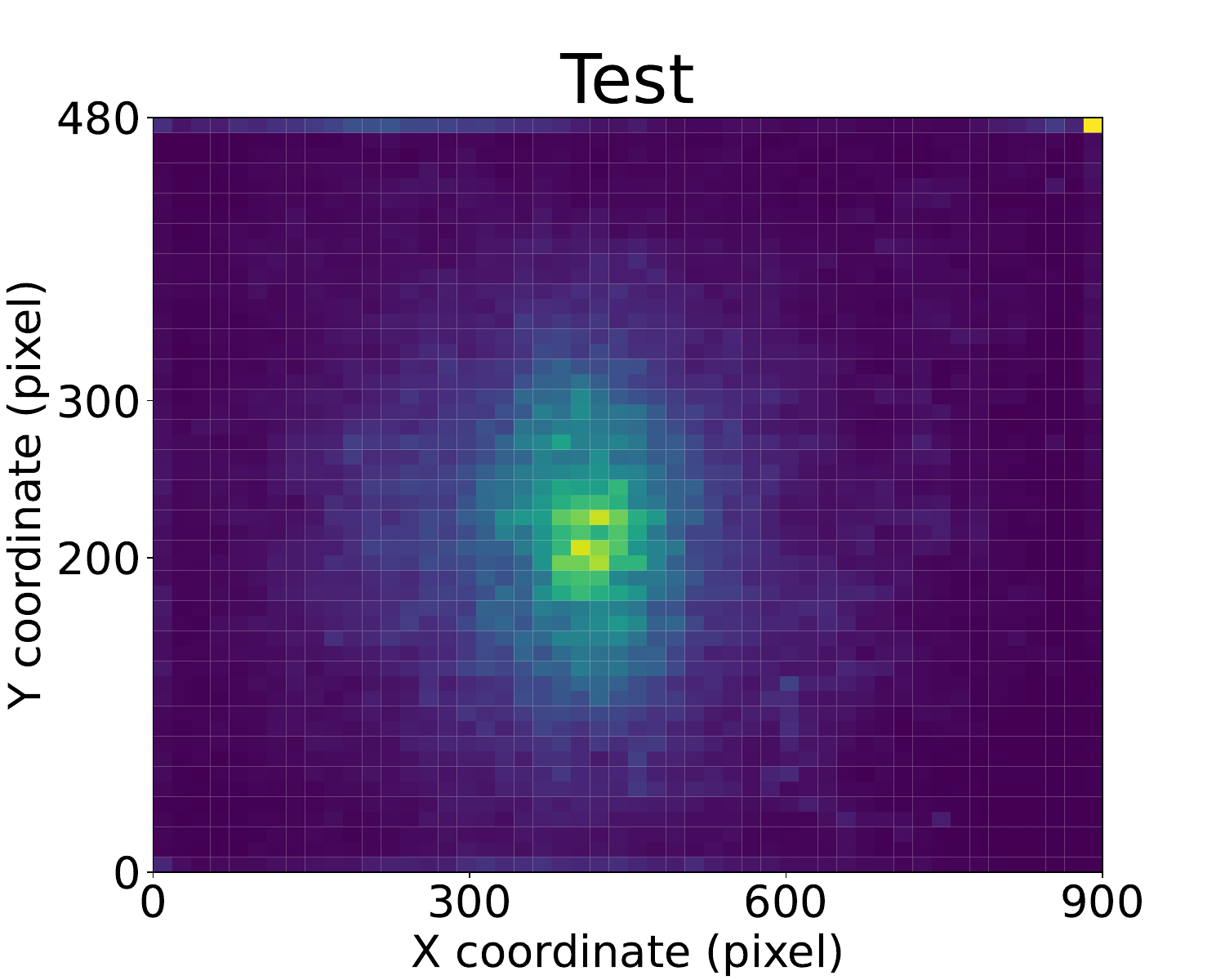}
\includegraphics[width=0.32\linewidth]{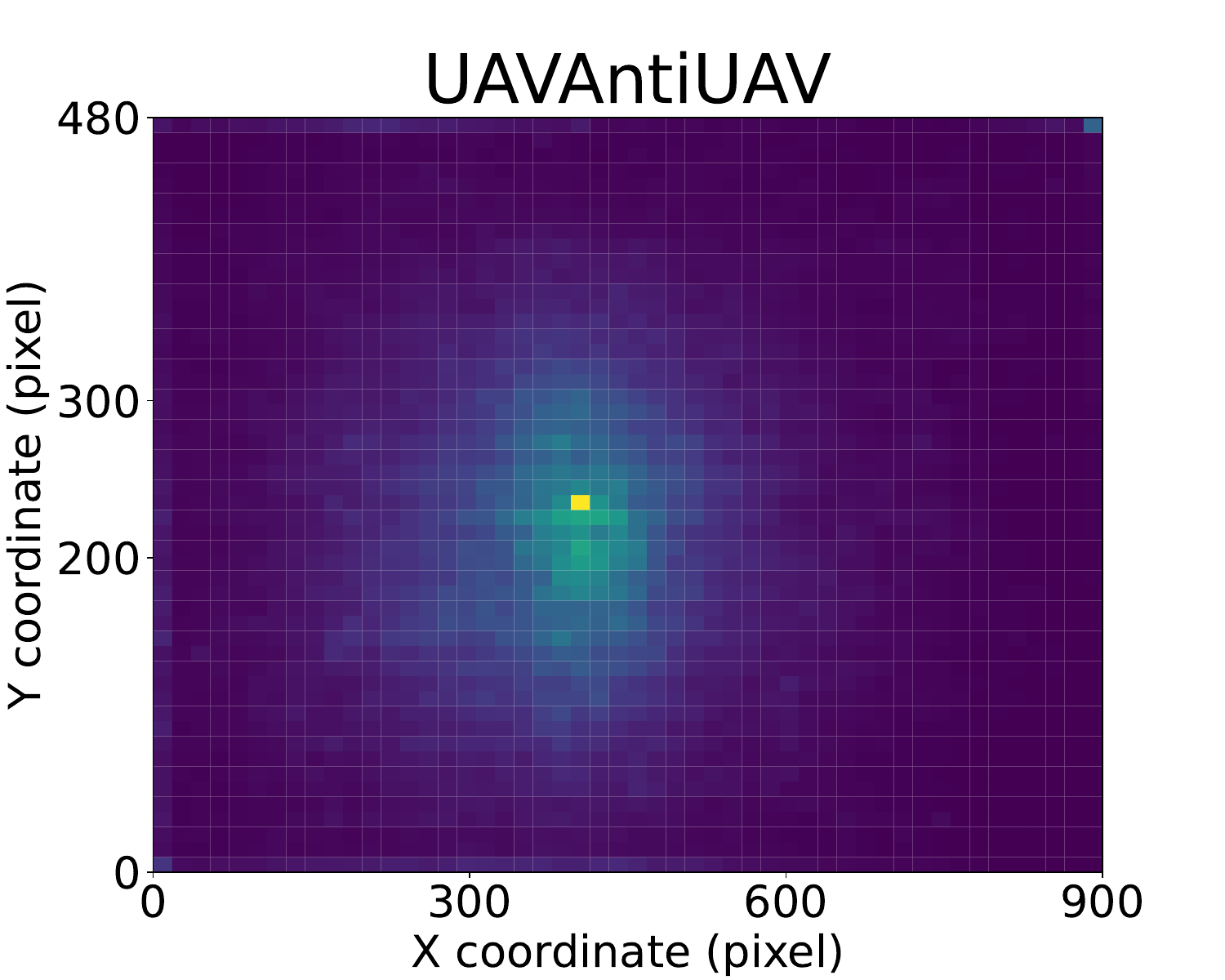}
\caption*{\color{black}(a) Position distribution}
\label{fig:position}
\end{minipage}
\begin{minipage}[t]{0.235\linewidth}
\centering
\includegraphics[width=1.0\linewidth]{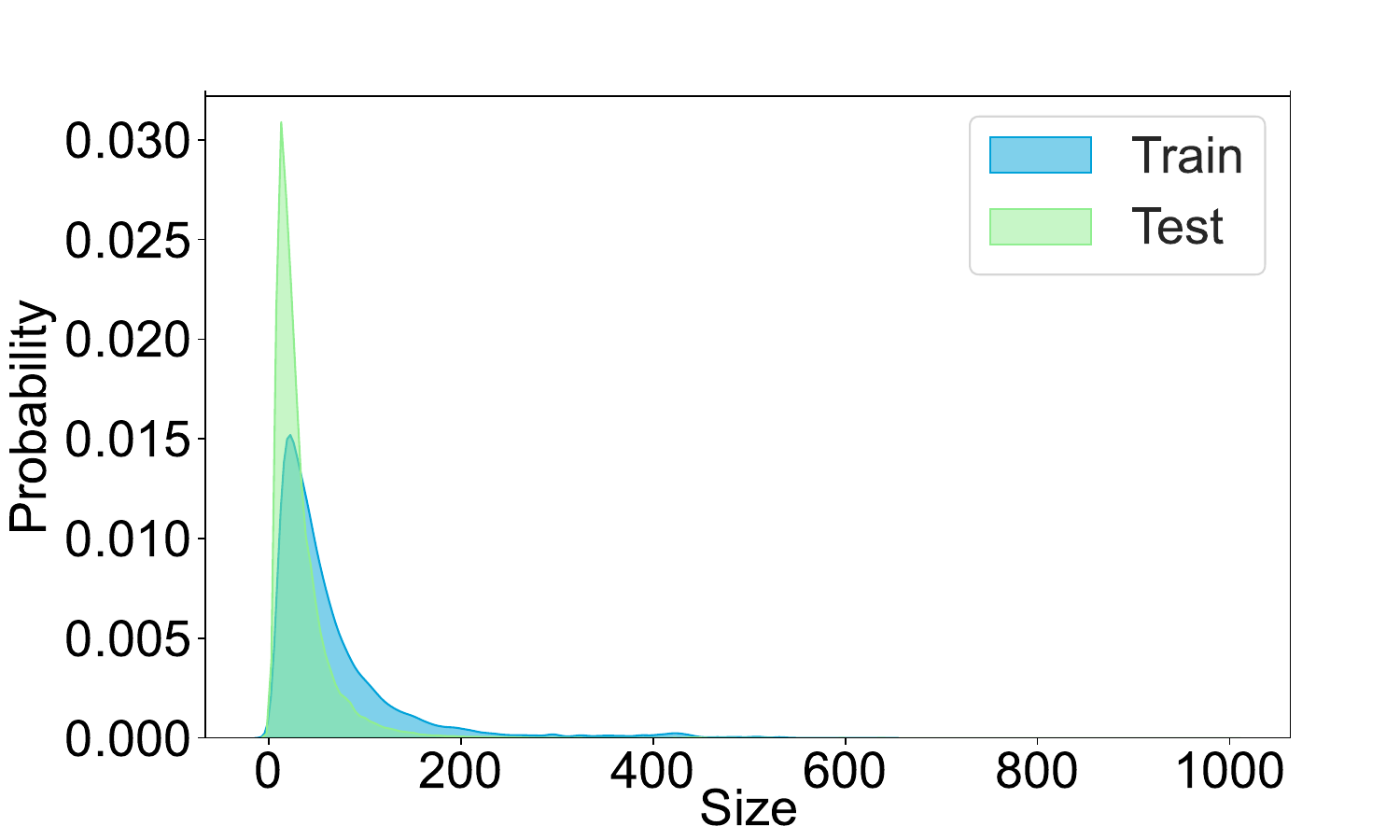}
\caption*{(b) Size distribution}
\label{fig:scale}
\end{minipage}
\begin{minipage}[t]{0.215\linewidth}
\centering
\includegraphics[width=1.0\linewidth]{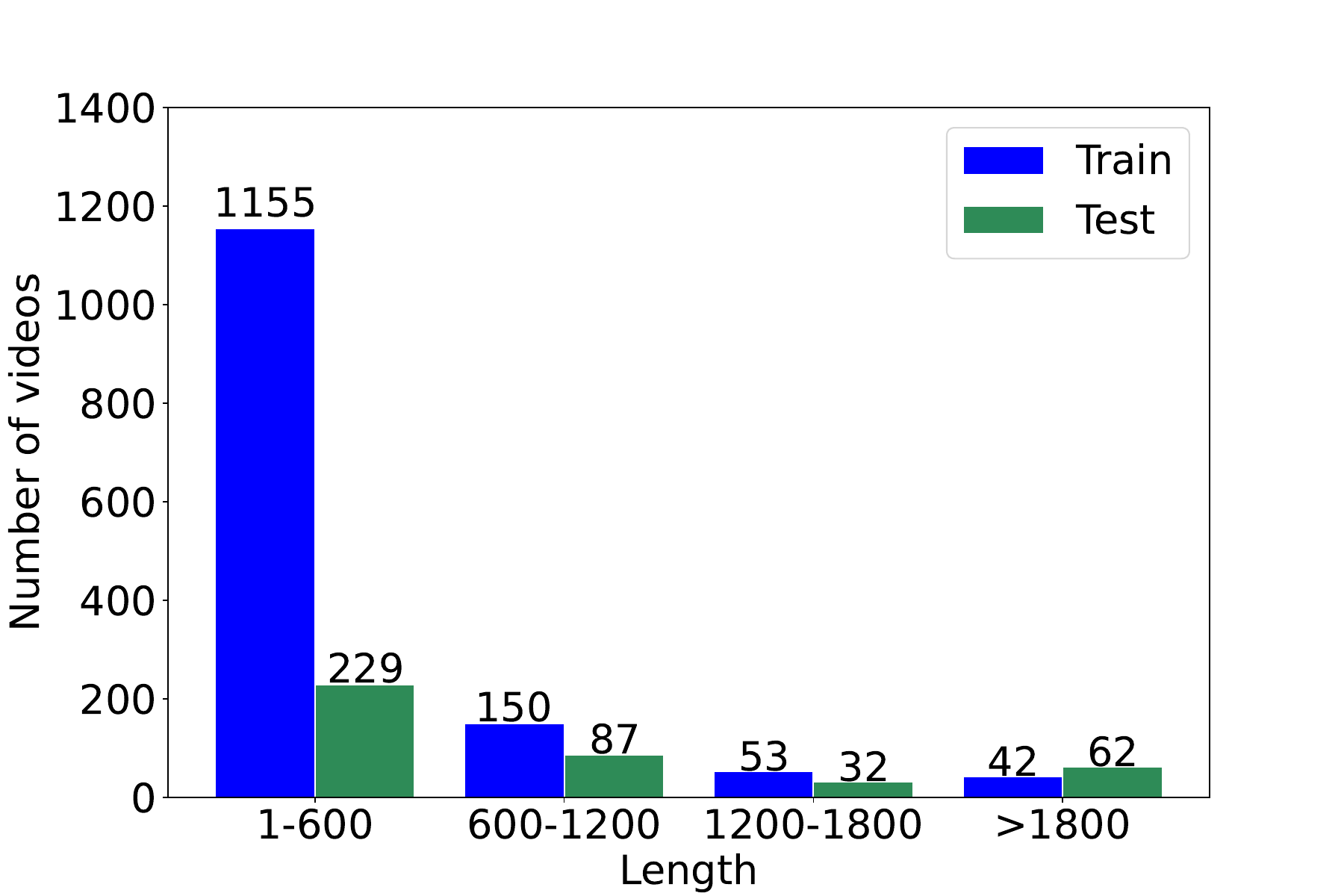}
\caption*{(c) Length distribution}
\label{fig:length}
\end{minipage}
\caption{Target position, size, and video length distributions in the UAV-Anti-UAV dataset. Best viewed by zooming in.}
\label{fig:statistical_analysis}
\end{figure*}

\section{UAV-Anti-UAV Dataset}
\label{sec:dataset}

To fuel research in UAV-Anti-UAV tracking, we construct a large-scale, meticulously annotated benchmark. An overview comparison with existing datasets is presented in Tab.~\ref{tab:Comp_UAVAntiUAV}.

\subsection{Data Collection}

Our data collection encompasses a wide variety of scenarios to ensure realism and diversity. We collect videos containing five distinct categories of UAVs as targets: fixed-wing drone, multi-rotor drone, vertical take-off and landing (VTOL)\footnote{As a subclass of VTOL, electric Vertical Take-Off and Landing (eVTOL) aircraft is evolving as a pivotal class of manned aerial vehicles, serving as the core of the low-altitude economy.} drone, first-person view (FPV) drone, and unmanned helicopters. Examples of each category with their language prompts are shown in Fig.~\ref{fig:examples}. The sequences were recorded in diverse environments (\eg, urban, rural, coastal, mountainous) under varying weather and lighting conditions. The final collection totals 1,810 video sequences, amounting to 1.05 million frames with a total duration of 9.85 hours.

\subsection{Data Annotation}

Annotations are performed manually by 10 expert annotators from a professional data annotation company, with a multi-stage review process to ensure accuracy. Each video is annotated with three complementary types of information to support comprehensive research: (1) Bounding Boxes: The target UAV is meticulously annotated with an axis-aligned bounding box in every frame. Specifically, the ground truth is denoted as $[x, y, w, h]$, where $(x, y)$ represents the top-left coordinates, and $w, h$ denote the width and height, respectively. (2) Language Prompts: Each sequence is accompanied by a concise, natural language description of the category of target UAV, its color, behavior, attributes, and surroundings (see Fig.~\ref{fig:examples} for examples). This enables research in vision-language tracking. (3) Tracking Attributes: To facilitate detailed performance diagnosis, we annotate each sequence with 15 challenging attributes.

\subsection{Attribute Definition} 

To enable in-depth evaluation of trackers, we label each video sequence with rich attributes. Specifically, we provide 15 attributes, including camera motion (CM), viewpoint changes (VC), partial occlusion (PO), full occlusion (FO), out-of-view (OV), rotation (ROT), similar distractors (SD), illumination variations (IV), motion blur (MB), partial target information (PTI), small object (SO), fast motion (FM), scale variations (SV), aspect ratio variations (ARV), and length (LEN) of video. Among them, following~\cite{zhang2025cost,zhu2023tiny}, we define a small object as a target with an average relative size smaller than $1\%$ of the video frame and an average absolute size less than $\sqrt{22 \times 22}$ pixels. The detailed definition of 15 tracking attributes is shown in Tab.~\ref{tab:Attribute_defination}.

\subsection{Statistical Analysis}

\myPara{Attribute Distribution.} As illustrated in Fig.~\ref{fig:attribute_of_videos}, the dataset presents a highly complex distribution of attributes that characterize the inherent difficulty of the UAV-Anti-UAV task. The dominant presence of Scale Variation (SV) and Aspect Ratio Variation (ARV), occurring in 1,324 and 1,040 sequences, respectively, coupled with Fast Motion (FM) in 1,018 sequences, underscores the severe dual-dynamic nature of the benchmark, where the target exhibits rapid and erratic state transitions relative to the observer. Furthermore, the substantial frequency of environmental challenges, including Partial Occlusion (PO, 861), Similar Distractors (SD, 602), and Out-of-View (OV, 567), suggests that the dataset imposes significant demands on target re-identification and discrimination capabilities in clutter. Finally, the diverse temporal distribution—comprising 1,384 short, 322 medium, and 104 long sequences—ensures a comprehensive evaluation spanning from short-term to long-term tracking.

\myPara{Brightness Distribution.} The brightness distribution presented in Fig.~\ref{fig:brightness_comparison} indicates that the proposed UAV-Anti-UAV dataset possesses a balanced and diverse illumination variations. With an average brightness value of 114.06, it is positioned centrally among existing benchmarks, bridging the gap between high-exposure datasets like DUT Anti-UAV (129.49) and extremely low-light nighttime UAV tracking datasets such as UAVDark135 (16.06). This intermediate mean value suggests that the dataset avoids bias toward specific lighting conditions, thereby offering a realistic spread of illumination scenarios that encompasses both bright daytime environments and dimmer settings.

\myPara{Relative Speed Distribution.} 
The relative speed~\cite{zhang2025cost} distribution presented in Fig.~\ref{fig:average_relative_speed_distribution} demonstrates that the proposed UAV-Anti-UAV dataset exhibits the highest motion intensity among all evaluated benchmarks. The dataset records an average relative speed of 0.79, which surpasses existing Anti-UAV datasets such as Anti-UAV318 (0.72) and DUT Anti-UAV (0.68). Furthermore, it significantly exceeds traditional UAV tracking datasets like UAV123 and UAV20L, which have average speeds of 0.46 and 0.41, respectively. Unlike the sharp peaks at lower speeds observed in other datasets, the UAV-Anti-UAV distribution shows a flatter curve with a heavier tail, reflecting the severe dual-dynamic disturbances and high-speed maneuvers inherent in air-to-air tracking scenarios.

\myPara{Position, Size, and Length Distributions.} Fig.~\ref{fig:statistical_analysis} reveals the dataset's diversity in target states and sequence durations. The position heatmaps in Fig.~\ref{fig:statistical_analysis}(a) demonstrate that while targets appear across the entire frame, there is a distinct concentration toward the center, a pattern characteristic of active pursuit scenarios where the pursuer aims to keep the target in view. Regarding target size, Fig.~\ref{fig:statistical_analysis}(b) highlights a highly varied distribution with a significant proportion of tiny objects (occupying less than 1\% of the image area), posing substantial challenges for precise localization and feature extraction. Furthermore, the video length distribution in Fig.~\ref{fig:statistical_analysis}(c) is long-tail, comprising 76.5\% short, 17.8\% medium, and 5.7\% long sequences. This temporal diversity ensures that trackers are rigorously evaluated on both short-term responsiveness and long-term robustness against error accumulation.

\begin{figure*}[ht]
  \centering
\includegraphics[width=1.0\linewidth]{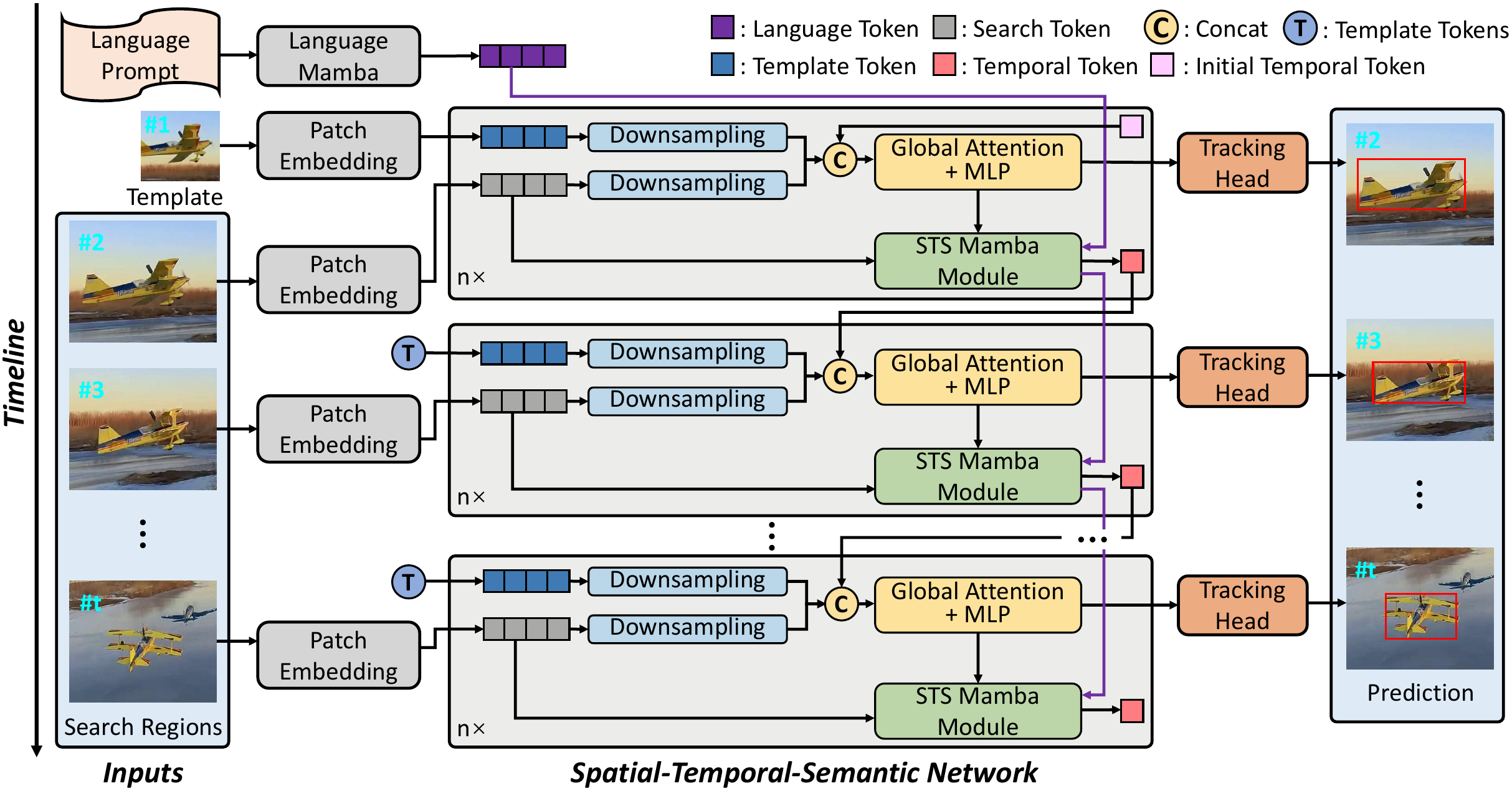}
  \caption{Overview of the proposed MambaSTS framework. The architecture accepts multi-modal inputs (template, search region, and language prompt) and processes them through a hierarchical spatial-temporal-semantic network. This network fuses visual and semantic features using the STS Mamba module, which leverages a temporal token propagation mechanism to capture video-level long-term context. Finally, an anchor-free tracking head predicts the target's state directly from the enhanced representations.}
  \label{fig:method}
\end{figure*}

\section{Methodology: MambaSTS}
\label{sec:methidology}

\subsection{Framework Overview}

To establish a strong baseline for the UAV-Anti-UAV task, we propose MambaSTS, a novel framework designed for integrated spatial-temporal-semantic learning. The overall architecture is depicted in Fig.~\ref{fig:method}. MambaSTS takes a template image $Z$, a search image $X$, and a language prompt $L$ as input. It processes these inputs through a multi-stream architecture where a hierarchical HiViT backbone~\cite{zhang2023hivit} extracts multi-scale visual features from the template and search regions, while a pre-trained language Mamba model extracts semantic tokens from the language prompt. These multi-modal features are then deeply fused in the spatial-temporal-semantic (STS) network, which stacks the STS Mamba module to model global spatial dependencies via attention mechanisms and long-term temporal context via a temporal token propagation mechanism that compresses historical target cues into a latent state. Finally, the enhanced feature representations are fed into a concise, anchor-free tracking head that predicts the target's classification, offset, and bounding box size directly from the search region tokens.

\subsection{Preliminaries}
In this subsection, we revisit the theoretical foundations of state space models (SSMs) and describe how they are adapted to process 2-D visual data. Therefore, we can employ SSMs, such as Mamba~\cite{gu2023mamba}, for both 1-D language and 2-D visual feature learning in our tracking framework.

\myPara{State Space Models.}
Modern SSM-based architectures, such as Mamba, are rooted in continuous linear time-invariant systems. Fundamentally, these systems map a 1-D input signal \(x(t) \in \mathbb{R}\) to an output response \(y(t) \in \mathbb{R}\) via an intermediate latent state \(h(t) \in \mathbb{R}^N\). This process is described by the following linear ordinary differential equation:
\begin{align}
    h'(t) &= \mathbf{A}h(t) + \mathbf{B}x(t), \\
    y(t) &= \mathbf{C}h(t),
\end{align}
where \(\mathbf{A} \in \mathbb{R}^{N \times N}\) denotes the state evolution matrix, while \(\mathbf{B} \in \mathbb{R}^{N \times 1}\) and \(\mathbf{C} \in \mathbb{R}^{1 \times N}\) serve as the projection parameters for the input and the state, respectively.

To integrate this continuous system into deep learning algorithms that operate on discrete sequences, a discretization process is essential. A timescale parameter \(\Delta\) is introduced to transform the continuous parameters \((\mathbf{A}, \mathbf{B})\) into their discrete counterparts \((\overline{\mathbf{A}}, \overline{\mathbf{B}})\). A standard approach for this transformation is the zero-order hold method, defined as:
\begin{align}
    \overline{\mathbf{A}} &= \exp(\Delta \mathbf{A}), \\
    \overline{\mathbf{B}} &= (\Delta \mathbf{A})^{-1}\left(\exp(\Delta \mathbf{A}) - \mathbf{I}\right) \cdot \Delta \mathbf{B}.
\end{align}

Consequently, the discretized system can be expressed recursively, enabling efficient autoregressive inference:
\begin{align}
    h_t &= \overline{\mathbf{A}}h_{t-1} + \overline{\mathbf{B}}x_t, \\
    y_t &= \mathbf{C}h_t.
\end{align}

Furthermore, to facilitate efficient parallel training on modern hardware, the operation can be reformulated as a global convolution as follows:
\begin{align}
    \overline{\mathbf{K}} &= \left(\mathbf{C}\overline{\mathbf{B}}, \mathbf{C}\overline{\mathbf{A}}\overline{\mathbf{B}}, \dots, \mathbf{C}\overline{\mathbf{A}}^{L_{s}-1}\overline{\mathbf{B}}\right), \\
    \mathbf{y} &= \mathbf{x} * \overline{\mathbf{K}},
\end{align}
where \(L_{s}\) represents the length of the input sequence and \(\overline{\mathbf{K}}\) is the structured convolutional kernel. Mamba further enhances this by making the parameters data-dependent, allowing the model to selectively propagate or suppress information along the sequence.

\myPara{Visual State Space Modeling.}
Standard Mamba blocks are inherently designed for 1-D sequential data. To adapt this architecture for visual tracking tasks, 2-D images must be tokenized into a sequence. Given an input image \(I \in \mathbb{R}^{H \times W \times C}\), we decompose it into flattened 2-D patches \(x_p \in \mathbb{R}^{J \times (P^2 \cdot C)}\), where \(P\) is the patch size, \(C\) is the channel dimension, and \(J\) is the total number of patches.

These patches are linearly projected into vectors of dimension \(D\). To preserve spatial information, which is critical for object localization, learnable position embeddings \(\mathbf{E}_{\text{pos}}\) are added to the patch sequence. Following the design of Vision Mamba (Vim)~\cite{zhu2024vision}, a learnable class token \(t_{\text{cls}}\) is appended to aggregate global information. The resulting input sequence \(T_0\) is formulated as follows:

\begin{equation}
    T_{0} = \left[t_{\text{cls}}; t_{p}^{1}\mathbf{W}; t_{p}^{2}\mathbf{W}; \dots; t_{p}^{J}\mathbf{W}\right] + \mathbf{E}_{\text{pos}},
\end{equation}
where \(\mathbf{W}\) is the learnable projection matrix and \(t_{p}^{j}\) represents the \(j\)-th image patch. Stacked Mamba block layers then process this tokenized sequence to extract robust feature representations.

\subsection{Visual and Language Representations}

Recent research indicates that Transformer models excel in visual representation learning but are limited by quadratic complexity in long contexts~\cite{chen2021transformer,wang2021transformer}. Conversely, Mamba models~\cite{gu2023mamba} offer efficient long-range modeling but struggle with non-autoregressive appearance features~\cite{yu2025mambaout,li2025mambalct}. To combine their strengths, we employ the Mamba model for global language features and the Transformer model for visual features. This hybrid strategy, illustrated in Fig.~\ref{fig:method}, enhances both semantic and spatial perception.

\myPara{Visual Feature Extraction.} To capture rich spatial information and adapt to the scale variations in aerial tracking, we employ HiViT~\cite{zhang2023hivit} as the visual backbone. As shown in Fig.~\ref{fig:method}, the visual backbone mainly contains the downsampling, global attention, and MLP layers. First, the initial template image $Z \in \mathbb{R}^{3 \times H_z \times W_z}$ and the current video search frame $X \in \mathbb{R}^{3 \times H_x \times W_x}$ are fed into the pipeline sequentially, where $H$ and $W$ denote the height and width, respectively. Unlike the standard ViT~\cite{dosovitskiy2020image}, which directly projects images into $16 \times 16$ flat patches, we adopt a hierarchical downsampling strategy to generate robust feature tokens. Specifically, the template $Z$ and search image $X$ pass through a three-stage process consisting of a $4 \times 4$ patch embedding layer followed by two consecutive $2 \times 2$ downsampling merging layers. This structure effectively transforms the inputs into compact visual tokens $F_z \in \mathbb{R}^{N_z \times D}$ and $F_x \in \mathbb{R}^{N_x \times D}$ with a channel dimension $D=512$. The resulting sequence lengths are $N_z = H_z W_z / P^2$ and $N_x = H_x W_x / P^2$, corresponding to a total stride of $P=16$, which balances computational efficiency with spatial resolution.

\myPara{Language Feature Extraction.} To introduce semantic guidance into the tracking process, we utilize a pre-trained language Mamba model~\cite{gu2023mamba} to process the language prompt $L$ (as shown in the top branch of Fig.~\ref{fig:method}). The input language prompt is first tokenized and converted into language tokens via the language Mamba tokenizer. Subsequently, the resulting sequence, prepended with a learnable \texttt{[CLS]} token, is fed into the language encoder to aggregate global semantic information, yielding the final language tokens $F_l \in \mathbb{R}^{N_l \times D}$. Here, we set the number of language tokens $N_l=40$.

The extracted visual tokens (\ie, template tokens \(F_z\) and search tokens \(F_x\)) and language tokens \(F_l\) are then fed into the STS Mamba module for unified spatial-temporal-semantic learning, providing comprehensive feature support for handling dual-dynamic disturbances and complex aerial tracking scenarios.

\subsection{Long-term Context via Temporal Token Propagation}
\label{sec:long_term_context}

In the UAV-Anti-UAV tracking scenario, the dual-dynamic nature involving rapid motion of both the pursuer and the target often leads to severe appearance deformation and frequent out-of-view situations. Recent research indicates that temporal context modeling is beneficial for mitigating variations in the target's appearance and motion states, thereby enhancing target perception capabilities of the tracking model~\cite{kang2025exploring,li2025mambalct}. However, relying solely on the current frame or short-term context (\eg, adjacent frames~\cite{xie2024autoregressive} or video clips~\cite{zheng2024odtrack}) is insufficient to maintain robust tracking when the target reappears after occlusion or blur. While Transformer-based methods~\cite{wang2021transformer,chen2021transformer} excel at spatial modeling, their quadratic computational complexity $\mathcal{O}(N^2)$ prohibits the processing of video-level long sequences. To bridge this gap, we leverage the linear complexity $\mathcal{O}(N)$ of the Mamba architecture~\cite{gu2023mamba} to establish a video-level long-term context.

\myPara{Temporal Token Propagation Mechanism.} Instead of treating tracking as isolated matching pairs, we formulate the long-term context modeling as a continuous sequence learning task. We propose a temporal token propagation mechanism that recursively compresses historical target cues into a compact latent state. Specifically, we construct a serialized feature stream from the search regions of all historical frames. Unlike standard approaches that discard historical features after processing, our module maintains a continuous flow of visual information.

Let $F = \{F_{x}^{1}, F_{x}^{2}, \dots, F_{x}^{i}\}$ denote the sequence of visual features extracted from the search regions from the first frame up to the current frame $i$. To effectively capture the target's evolutionary trajectory and appearance changes over time, we employ a selective state space model to scan this sequence. The construction of the long-term context is formulated as:
\begin{equation}
    \mathcal{T}_{temp}^i \leftarrow \text{Mamba} : \{F_{x}^{1}, F_{x}^{2}, F_{x}^{3}, \dots, F_{x}^{i}\},
    \label{eq:long_term_context}
\end{equation}
where $F_{x}^{i}$ represents the flattened visual tokens of the $i$-th search frame, and $\mathcal{T}_{temp}^i$ denotes the generated \textbf{temporal token} at the current frame $i$.

\myPara{Mechanism Implementation.} The initial temporal token is randomly initialized. As illustrated in the system pipeline (see Fig.~\ref{fig:method}), the temporal token $\mathcal{T}_{temp}^i$ acts as a carrier of ``video memory''. Through the Mamba model's selective scanning mechanism, the model dynamically filters out irrelevant background noise (\eg, clouds, ground clutter) while selectively retaining target-specific features from previous frames. This accumulated historical information is then propagated to the next frame's processing stage.

Mathematically, the recursion is governed by the discretized state space equation: $h_t = \bar{\mathbf{A}}h_{t-1} + \bar{\mathbf{B}}F^{t}$. The temporal token is essentially a projection of the current hidden state $h_t$, encapsulating the accumulated target variation cues from frame 1 to $i$. By injecting this temporal token into the visual backbone of the current frame (see Fig.~\ref{fig:method}), we provide the tracking model with a robust prior of the target's motion and appearance state. This ensures that even when the target undergoes full occlusion or motion blur, the tracker maintains a consistent internal representation derived from the complete historical context, enabling rapid re-acquisition.

\subsection{Unified Spatial-Temporal-Semantic Modeling}
\label{sec:unified_modeling}

To achieve deep interaction among visual appearance, semantic cues, and historical context, we construct a spatial-temporal-semantic network by integrating the proposed STS Mamba module into the hierarchical visual backbone. As illustrated in Fig.~\ref{fig:method}, this module is inserted at multiple stages of the HiViT~\cite{zhang2023hivit} backbone, enabling progressive fusion of multi-modal features at different resolutions.

\myPara{Multi-Modal Input Interaction.} Unlike previous unified modeling approaches that only consider visual streams, our framework explicitly incorporates semantic language tokens to guide the tracking process. At each stage $i$, we first align the spatial dimensions of the template features $F_{z}^{i}$ and search region features $F_{x}^{i}$ through a downsampling operation, denoted as $\text{Ds}(\cdot)$. These visual tokens are then concatenated with the language tokens $F_{l}$ and the historical temporal token $\mathcal{T}_{temp}$ (carrying motion priors).

The construction of the unified feature sequence $Z_{in}^i$ for the $i$-th stage is formulated as:
\begin{equation}
    \begin{aligned}
    \hat{F}_{z}^{i}, \hat{F}_{x}^{i} &= \text{Ds}(F_{z}^{i}, F_{x}^{i}), \\
    Z_{in}^i &= \text{Concat}(\hat{F}_{z}^{i}, \hat{F}_{x}^{i}, F_{l}, \mathcal{T}_{temp}),
    \end{aligned}
    \label{eq:input_fusion}
\end{equation}
where $\text{Concat}(\cdot)$ denotes concatenation along the token dimension. This unified sequence $Z_{in}^i$ encapsulates the spatial detail, semantic intent, and temporal history required for robust target estimation.

\myPara{STS Mamba Module.} The core engine for processing $Z_{in}^i$ is the STS Mamba module. We adopt the architecture of Vision Mamba (Vim)~\cite{zhu2024vision} with $m$ blocks as the foundation due to its efficiency in modeling long sequences. However, the standard Vim employs a bidirectional scanning strategy (forward and backward) to capture global context for static images. In the tracking scenario, the target's state is strictly causal--current estimation depends only on historical information.

Therefore, we make a critical modification to the Vim block by employing a unidirectional scanning mechanism. Specifically, for the fused sequence $Z_{in}^i$, the state space modeling is constrained to propagate information only in the forward temporal direction. This ensures that the generated features strictly adhere to the causal dependency of the video stream. The modeling process of the STS Mamba module can be expressed as:
\begin{equation}
    \begin{aligned}
    H_k &= \bar{\mathbf{A}}H_{k-1} + \bar{\mathbf{B}}Z_{in}^i[k], \\
    Y_k &= \mathbf{C}H_k, \\
    Z_{out}^i &= \text{Linear}(Y_{K}) + Z_{in}^i,
    \end{aligned}
    \label{eq:sts_mamba}
\end{equation}
where $H_k$ represents the hidden state at step $k$, and $\bar{\mathbf{A}}, \bar{\mathbf{B}}, \mathbf{C}$ are the discretized SSM parameters derived from the unidirectional scan. $Y_{K}$ aggregates video-level long-term context information through the hidden state, which is used to update the temporal token $\mathcal{T}_{temp}$.

\myPara{Hierarchical Stacking.} As shown in the spatial-temporal-semantic network of Fig.~\ref{fig:method}, these STS Mamba modules are integrated into the hierarchical visual backbone  HiViT. In each stage, the output $F_{out}^i$ is split back into its constituent components. The updated visual tokens are fed into the subsequent HiViT layers (global attention + MLP) for spatial refinement, while the updated temporal token is propagated to the next stage. This layer-wise interaction ensures that semantic and temporal contexts deeply modulate the visual feature representation from coarse to fine levels, significantly enhancing the tracker's ability to distinguish the target object from distractors under dual-dynamic disturbances.

\subsection{Tracking Head and Loss Function}
\label{sec:tracking_head}

To translate the learned spatial-temporal-semantic representations into precise target states, we employ a concise fully convolutional network as the tracking head. Specifically, only the feature tokens corresponding to the search region are utilized for prediction. These tokens are reshaped back into a 2D feature map $F_{search} \in \mathbb{R}^{\frac{H_{x}}{P} \times \frac{W_{x}}{P} \times D}$ and fed into the prediction module.

\myPara{Tracking Head.} Following the paradigm of anchor-free tracking, the head comprises three parallel branches to predict classification scores and bounding box geometries:
\begin{itemize}
    \item \textbf{Classification Branch:} Outputs a score map $\mathbf{P} \in \mathbb{R}^{\frac{H_{x}}{P} \times \frac{W_{x}}{P} \times 1}$ indicating the probability of the target center's presence at each grid location.
    \item \textbf{Offset Branch:} Outputs a local offset map $\mathbf{O} \in \mathbb{R}^{\frac{H_{x}}{P} \times \frac{W_{x}}{P} \times 2}$ to compensate for the discretization error caused by the patch embedding process.
    \item \textbf{Bounding Box Size Branch:} Outputs a size map $\mathbf{B} \in \mathbb{R}^{\frac{H_{x}}{P} \times \frac{W_{x}}{P} \times 2}$ estimating the width and height of the target.
\end{itemize}

\myPara{Loss Function.} The model is trained using a linear combination of classification and regression losses. We utilize the weighted focal loss ($\mathcal{L}_{cls}$)~\cite{law2018cornernet} for the classification branch to handle the imbalance between positive and negative samples. For bounding box regression, we combine the $L_1$ loss ($\mathcal{L}_{1}$) and the generalized IoU loss ($\mathcal{L}_{GIoU}$)~\cite{rezatofighi2019generalized} to enforce precise localization. The total objective function is defined as:
\begin{equation}
    \mathcal{L}_{total} = \mathcal{L}_{cls} + \lambda_1 \mathcal{L}_{1} + \lambda_2 \mathcal{L}_{GIoU},
\end{equation}
where $\lambda_1$ and $\lambda_2$ are weighting factors set to 5 and 2, respectively, following standard conventions in visual tracking~\cite{ye2022joint}.

\begin{table}[ht]
\footnotesize
  \caption{Summary of 50 modern deep tracking algorithms. ``Trans.'' denotes Transformer. ``B'', ``S'', and ``T''  represent base, small, and tiny models.}
  \label{tab:summary_of_trackers}
  \centering
  \setlength{\tabcolsep}{0.01mm}{
  \scalebox{1.0}{
  \begin{tabular}{lcccc}
    \Xhline{0.75pt} 
    \textbf{Tracker}     & \textbf{Source }    & \textbf{Backbone }& \textbf{FPS}~ & ~\textbf{VL-based}\\
    \hline
    
    01. SiamFC~\cite{bertinetto2016fully} & ECCVW16  &   AlexNet  & 86    & \xmark \\
    
    02. ECO~\cite{danelljan2017eco}   &  CVPR17 &     VGG-M  & 8       & \xmark \\
    
    03. VITAL~\cite{song2018vital} &     CVPR18  &     VGG-M   &  1.5  & \xmark \\
    
    04. ATOM~\cite{danelljan2019atom} &     CVPR19  &     ResNet-18   & 30   & \xmark \\

    05. SiamMask~\cite{wang2019fast} &     CVPR19  &    ResNet-50   & 55   & \xmark \\
        
    06. SiamRPN++~\cite{li2019siamrpn++} &     CVPR19  &      ResNet-50   & 35   & \xmark \\

    07. SiamFC++~\cite{XuWLYY20} & AAAI20  &      GoogleNet   & 90   & \xmark \\
        
    08. SiamBAN~\cite{ChenZLZJ20} &    CVPR20  &     ResNet-50   & 40   & \xmark \\
    
    09. SiamCAR~\cite{guo2020siamcar} &    CVPR20  &     ResNet-50   & 52  & \xmark \\

    10. LightTrack~\cite{yan2021lighttrack}  & CVPR21 &    DSConv, MBConv   & 60   & \xmark \\

    11. SiamGAT~\cite{guo2021graph} & CVPR21 &      Inception v3   &   70  & \xmark \\
        
    12. TrDiMP~\cite{wang2021transformer} &     CVPR21  &      ResNet-50, Trans.   & 26   & \xmark \\  
    
    13. TransT~\cite{chen2021transformer} &    CVPR21  &     ResNet-50, Trans.   & 50   & \xmark \\
    
    14. STARK-ST50~\cite{yan2021learning} &     ICCV21  &     ResNet-50, Trans.   & 40   & \xmark \\

    15. KeepTrack~\cite{mayer2021learning} &    ICCV21  &     ResNet-50   & 18  & \xmark \\

    16. HiFT~\cite{cao2021hift} & ICCV21  & AlexNet, Trans. &   130 &  \xmark\\
    
    17. AutoMatch~\cite{zhang2021learn} &     ICCV21  &      ResNet-50  & 50   & \xmark \\

    18. TCTrack~\cite{cao2022tctrack} &    CVPR22  &     AlexNet   & 126   & \xmark \\
    
    19. ToMP-101~\cite{mayer2022transforming} &     CVPR22  &     ResNet-101, Trans.   & 25   & \xmark \\

    20. AiATrack~\cite{gao2022aiatrack} &    ECCV22  &     ResNet-50, Trans.   & 38   & \xmark \\
        
    21. SimTrack-B32~\cite{chen2022backbone} &     ECCV22  &     ViT-B   & 30   & \xmark \\
    
    22. OSTrack~\cite{ye2022joint} &     ECCV22  &     ViT-B   & 105   & \xmark \\
        
    23. ZoomTrack~\cite{kou2023zoomtrack} & NeurIPS23 &      ViT-B   & 100   & \xmark \\
    
    24. MixFormerV2-B~\cite{cui2023mixformerv2} &    NeurIPS23  &     ViT-B   & 165   & \xmark \\

    25. Aba-ViTrack~\cite{li2023adaptive} &    ICCV23  &     Aba-ViT   & 182  & \xmark \\

    26. GRM~\cite{gao2023generalized} &    CVPR23  &     ViT-B   & 45  & \xmark \\

    27. DropTrack~\cite{wu2023dropmae} & CVPR23 &      ViT-B   & 58   & \xmark \\
    
    28. SeqTrack-B256~\cite{chen2023seqtrack} &    CVPR23  &     ViT-B   & 40   & \xmark \\

    29. TCTrack++~\cite{cao2023towards} & TPAMI23 &      AlexNet   & 122   & \xmark \\

    30. ODTrack~\cite{zheng2024odtrack} & AAAI24  &      ViT-B   & 32   & \xmark \\

    31. EVPTrack~\cite{shi2024explicit} & AAAI24 &      HiViT-B   & 71   & \xmark \\
    
    32. AQATrack~\cite{xie2024autoregressive} & CVPR24 &      HiViT-B   & 65   & \xmark \\
    
    33. HIPTrack~\cite{cai2024hiptrack} & CVPR24 &      ViT-B   & 45   & \xmark \\

    34. ARTrackV2~\cite{bai2024artrackv2} & CVPR24 &      ViT-B   & 94   & \xmark \\

    35. LORAT-B224~\cite{lin2024tracking} & ECCV24 &      ViT-B   & 209   & \xmark \\

    36. MambaNUT~\cite{wu2024mambanut} & IROS25 &      Mamba-S   & 72   & \xmark \\

    37. SGLATrack~\cite{xue2025similarity} & CVPR25 &      DeiT-T    &  225   & \xmark \\
        
    38. ORTrack~\cite{wu2025learning} & CVPR25 &      DeiT-T   &  226   & \xmark \\
    
    39. MambaLCT~\cite{li2025mambalct} & AAAI25 &   Vim-S, HiViT-B   &  59   & \xmark \\

    40. MCITrack-B224~\cite{kang2025exploring} & AAAI25 &      Fast-iTPN-B, Mamba   & 35   & \xmark \\
        
    \hdashline    
    
    41. VLT$_{\rm TT}$~\cite{guo2022divert} &     NeurIPS22  &      ResNet-50, Bert-B   & 35   & \cmark \\
     
    42. JointNLT~\cite{zhou2023joint} &     CVPR23  &      Swin-B, Bert-B    & 39   & \cmark \\
    
    43. CiteTracker-256~\cite{li2023citetracker} &     ICCV23  &     ViT-B, CLIP-B   & 24   & \cmark \\
    
    44. All-in-One~\cite{zhang2023all} &    ACM MM23  &     ViT-B, Bert-B   & 60  & \cmark \\
     
    45. UVLTrack~\cite{ma2024unifying} &    AAAI24  &     ViT-B, Bert-B   & 57   & \cmark \\
    
    46. MambaTrack~\cite{zhang2025mambatrack} & ICASSP25 &    Mamba, Vim-S  & 42   & \cmark \\

    47. SUTrack-B224~\cite{chen2025sutrack} & AAAI25 &      HiViT-B, CLIP-L   & 55   & \cmark \\

    48. DUTrack-256~\cite{li2025dynamic} & CVPR25 &      HiViT-B, Bert-B    &   44   & \cmark \\

    49. ATCTrack~\cite{feng2025atctrack} & ICCV25 &     Fast-iTPN-B, RoBERTa-B   &    35   & \cmark \\
    
    \hdashline
    
    50. MambaSTS &    Ours  &     Mamba, Trans.   & 54   & \cmark \\
    \Xhline{0.75pt} 
  \end{tabular}
}}
\end{table}

\section{Experiments}
\label{sec:experiments}
Experiments are conducted on an Ubuntu 20.04 server with 8 NVIDIA A6000 GPUs (48GB memory), Intel Xeon 8375C CPU (32 cores), and 512GB RAM. Our MambaSTS is implemented using PyTorch 2.1.1 and achieves an inference speed of 54 FPS on a single NVIDIA A6000 GPU.

\subsection{Implementation Details}
\label{sec:implementation}

\myPara{Model Settings.}  For the visual branch, we utilize HiViT-base~\cite{zhang2023hivit} with $n=20$ blocks, initialized with MAE~\cite{he2022masked}, as the backbone. The STS Mamba module is constructed by stacking $m=24$ layers of Vim-small blocks~\cite{zhu2024vision}, which are adapted to perform unidirectional scanning for causal temporal modeling. For the language branch, we adopt GPT-NeoX~\cite{gpt-neox-library} and Mamba-130M~\cite{gu2023mamba} as the text tokenizer and language encoder, respectively. During data processing, the template and search regions are cropped to $2^2$ and $4^2$ times the target area and then resized to $128 \times 128$ and $256 \times 256$, respectively.

\myPara{Training.} We train our model on a combination of standard tracking datasets, including GOT-10k~\cite{huang2019got}, LaSOT~\cite{fan2019lasot}, COCO~\cite{lin2014microsoft}, and TrackingNet~\cite{muller2018trackingnet}, along with the training set of our proposed UAV-Anti-UAV dataset. Following~\cite{li2025mambalct}, we sample video clips with a length of 2 frames to learn temporal transitions. The model is optimized using the AdamW optimizer with a weight decay of $1 \times 10^{-4}$ and a batch size of 32. The initial learning rate is set to $2 \times 10^{-4}$ for the visual backbone and $2 \times 10^{-3}$ for the remaining parameters. The training process spans 300 epochs, with the learning rate decaying by a factor of 10 after the 240th epoch. 

\myPara{Inference.} During the inference phase, MambaSTS operates in a recursive manner. The search features and the temporal token from the previous step are fed into the STS Mamba module. The temporal token acts as a medium for information transmission, propagating video-level context frame-by-frame to maintain robust tracking without the need for online template updates. Since language features are extracted only once per video sequence, the associated computational overhead can be negligible.

\subsection{Evaluation Metrics}

We adopt the one-pass evaluation (OPE) protocol~\cite{wu2015otb,zhang2022webuav}, widely used in tracking benchmarks. Note that the OPE protocol evaluates trackers on the entire sequence without re-initialization, simulating real-world scenarios. We use five metrics to comprehensively measure performance, \ie, precision (Pre), normalized precision (nPre), success rate (AUC), complete success rate (cAUC), and mean accuracy (mACC).

\begin{figure*}[t]
\centering
\subfloat{\includegraphics[width =0.5\linewidth]{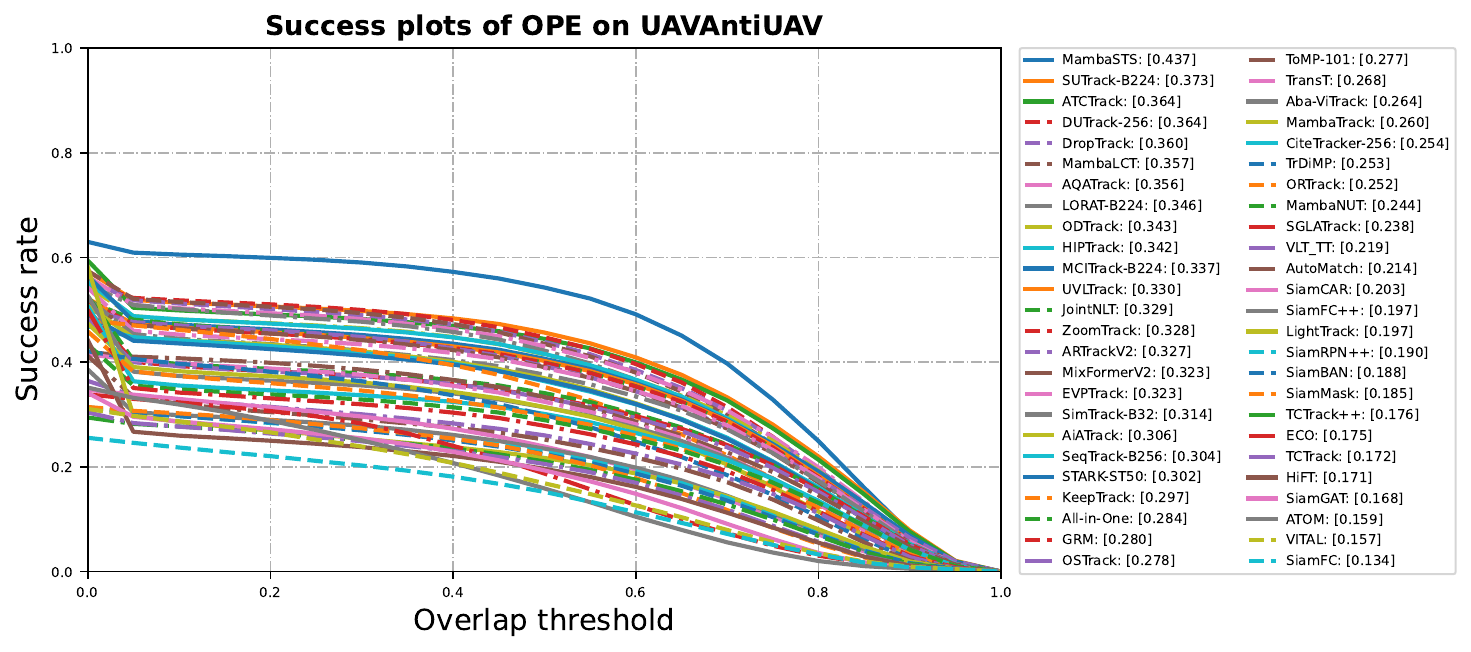}}
~\subfloat{\includegraphics[width =0.5\linewidth]{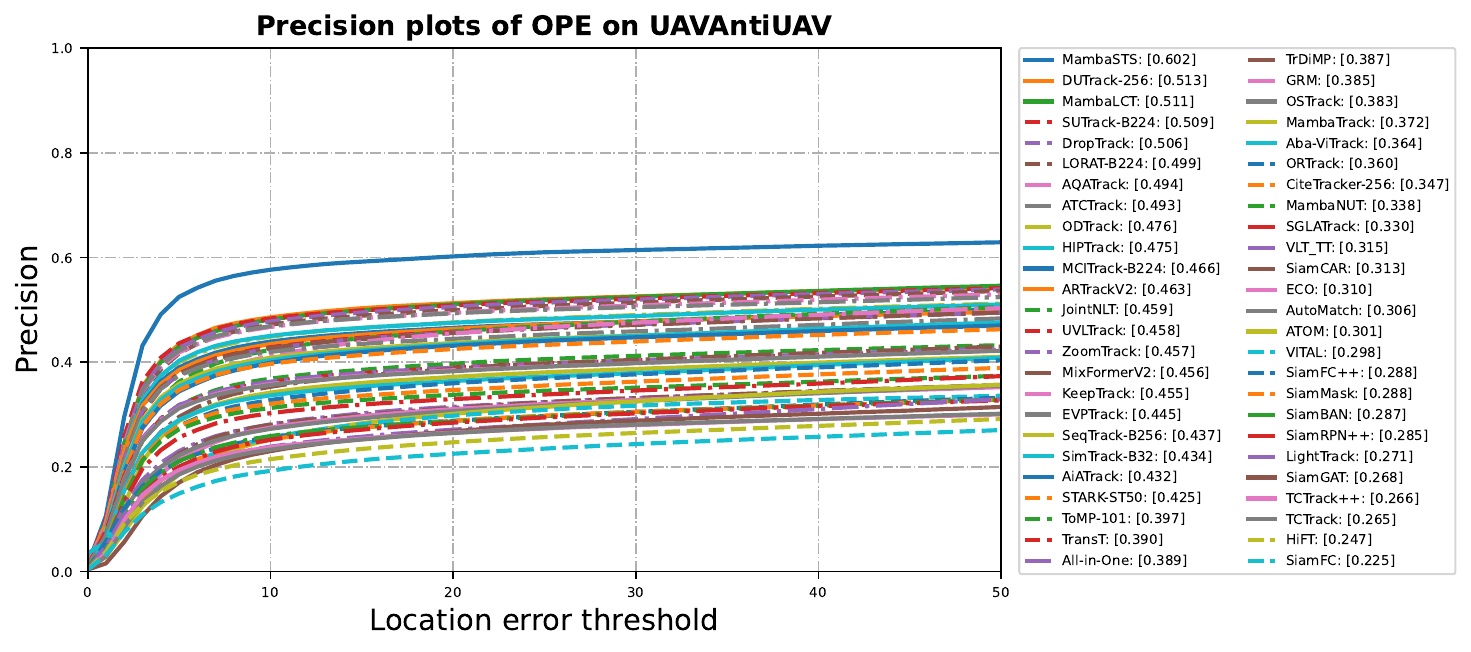}}\\
\subfloat{\includegraphics[width =0.5\linewidth]{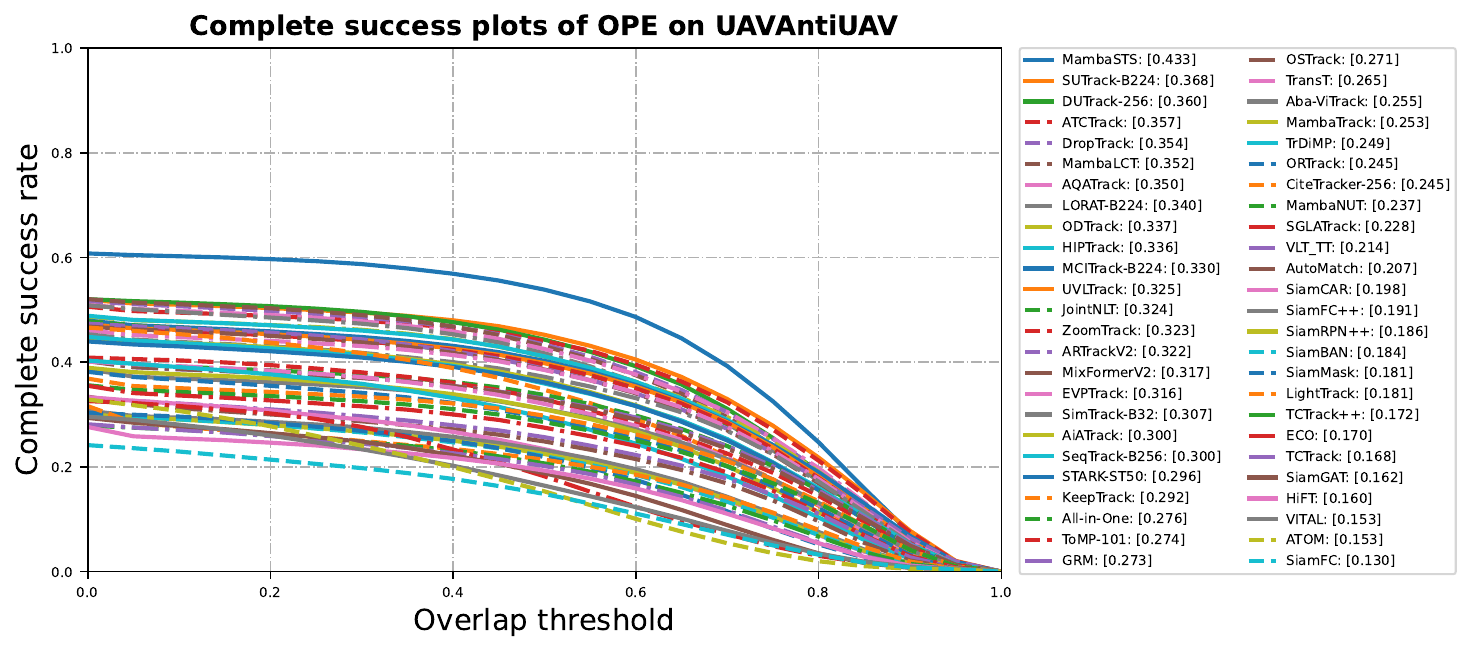}}
~\subfloat{\includegraphics[width =0.5\linewidth]{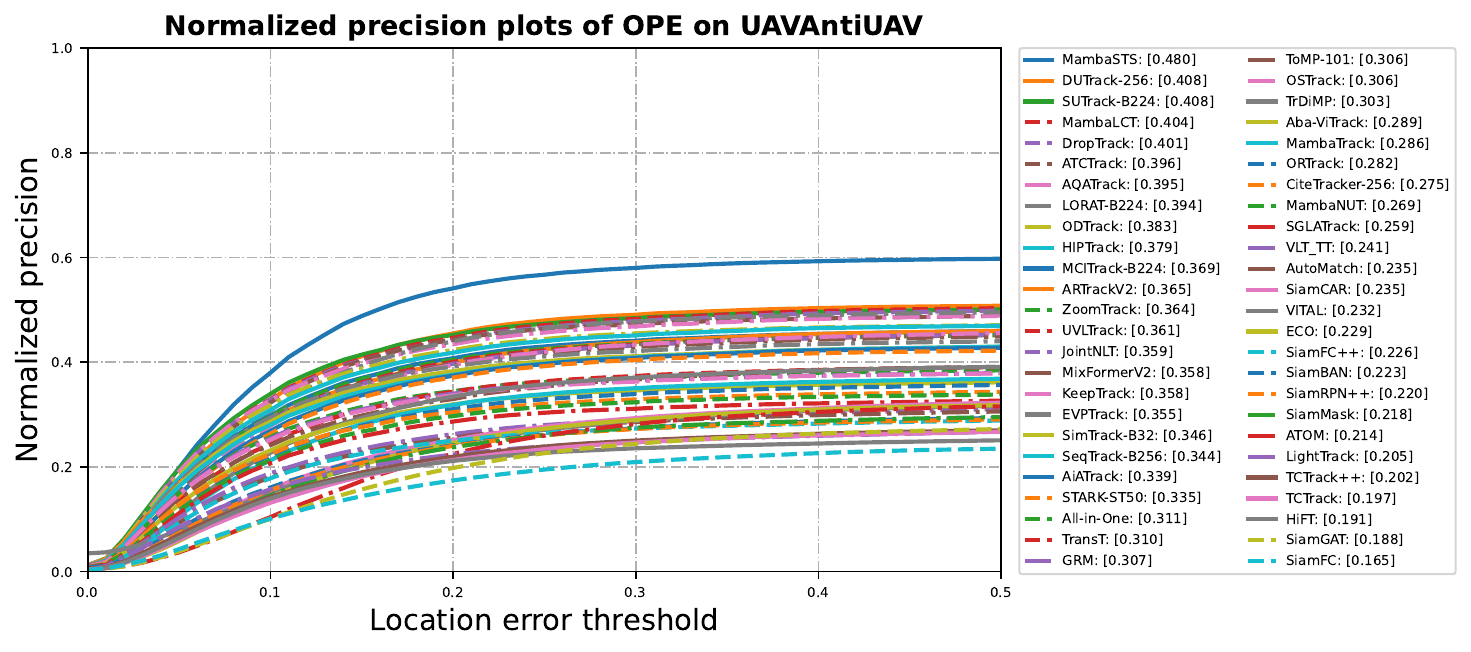}}
\caption{Benchmark results of 50 representative deep trackers on the UAV-Anti-UAV dataset using AUC, Pre, cAUC, and nPre scores. Best viewed in color with zooming in.}
\label{fig:overall_performance}
\end{figure*}

\begin{figure}[ht]
  \centering
\includegraphics[width=1.0\linewidth]{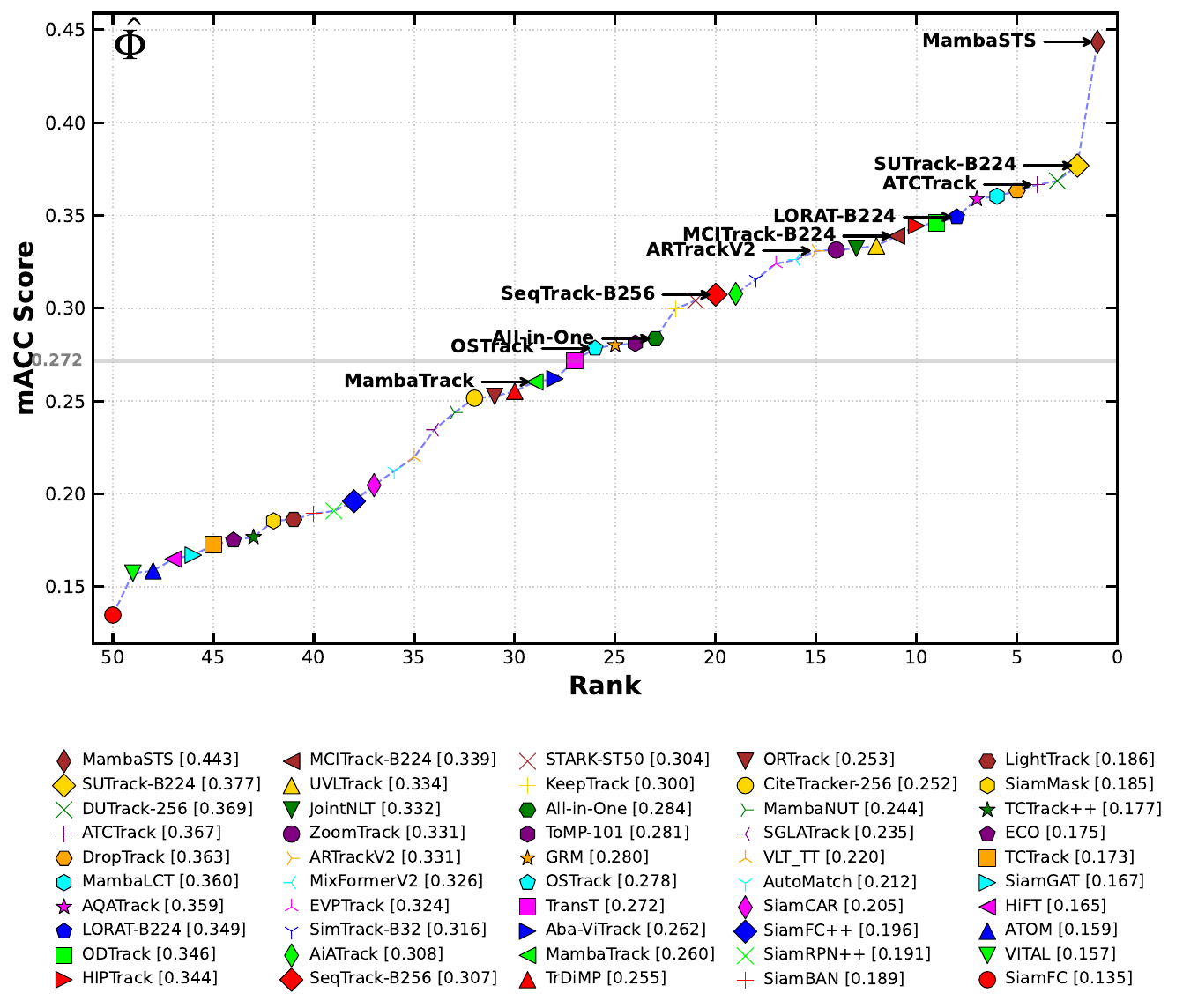}
  \caption{Comparison of mACC scores of 50 SOTA trackers on the UAV-Anti-UAV dataset. The gray horizontal line indicates
 the average performance of 50 trackers.}
  \label{fig:mACC}
\end{figure}

\begin{figure*}[ht]
\centering
\subfloat{\includegraphics[width =0.25\linewidth]{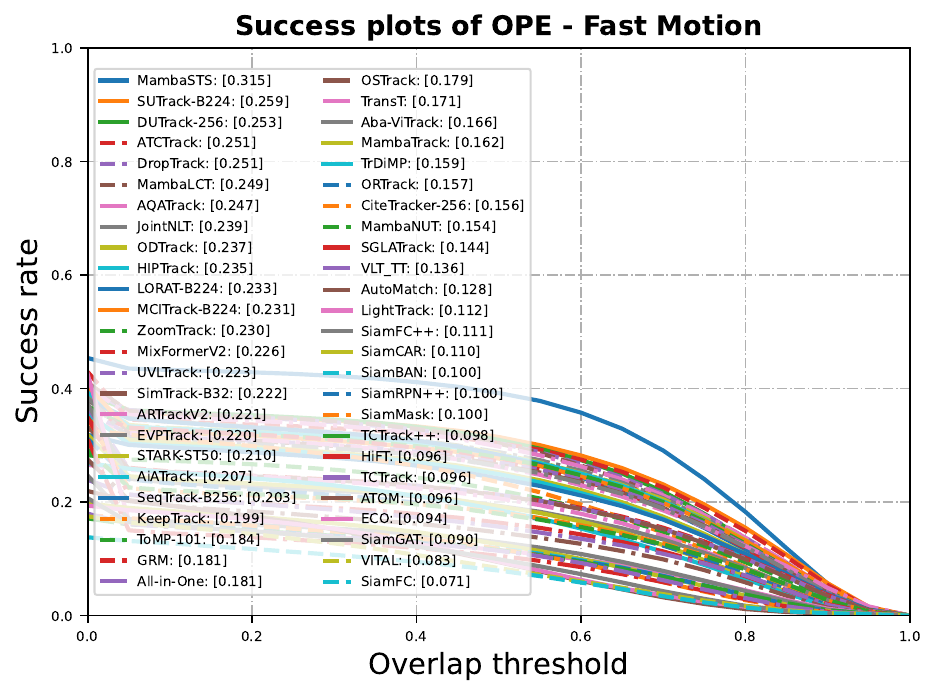}}
~\subfloat{\includegraphics[width =0.25\linewidth]{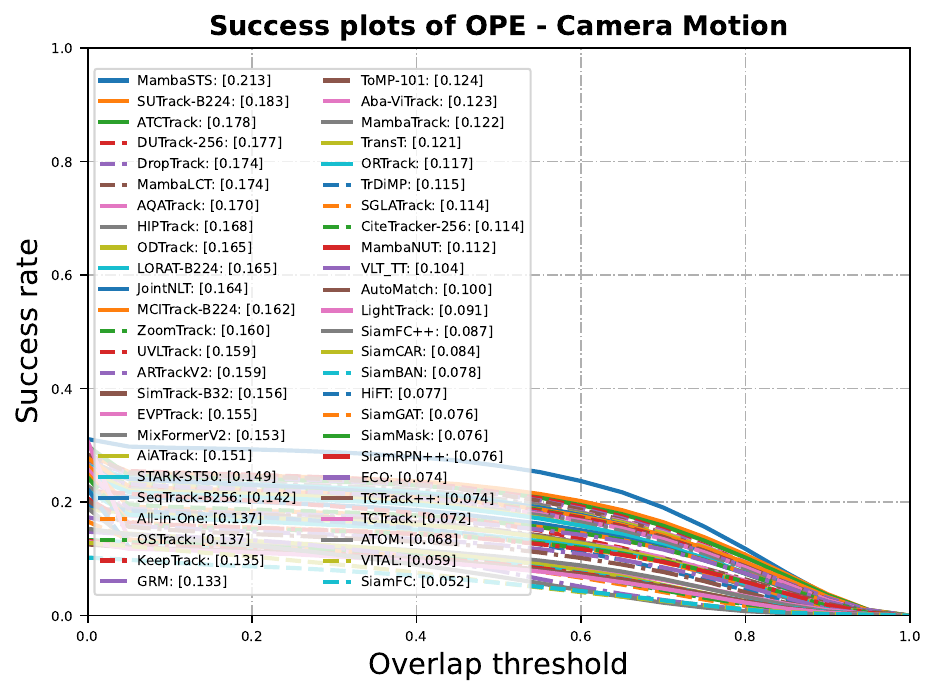}}
\subfloat{\includegraphics[width =0.25\linewidth]{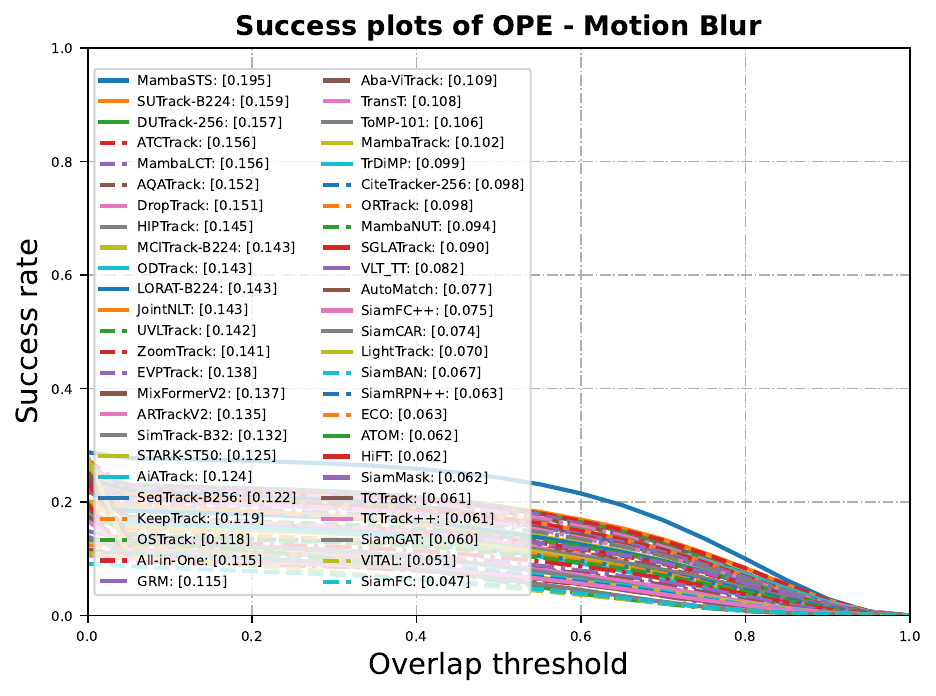}} 
~\subfloat{\includegraphics[width =0.25\linewidth]{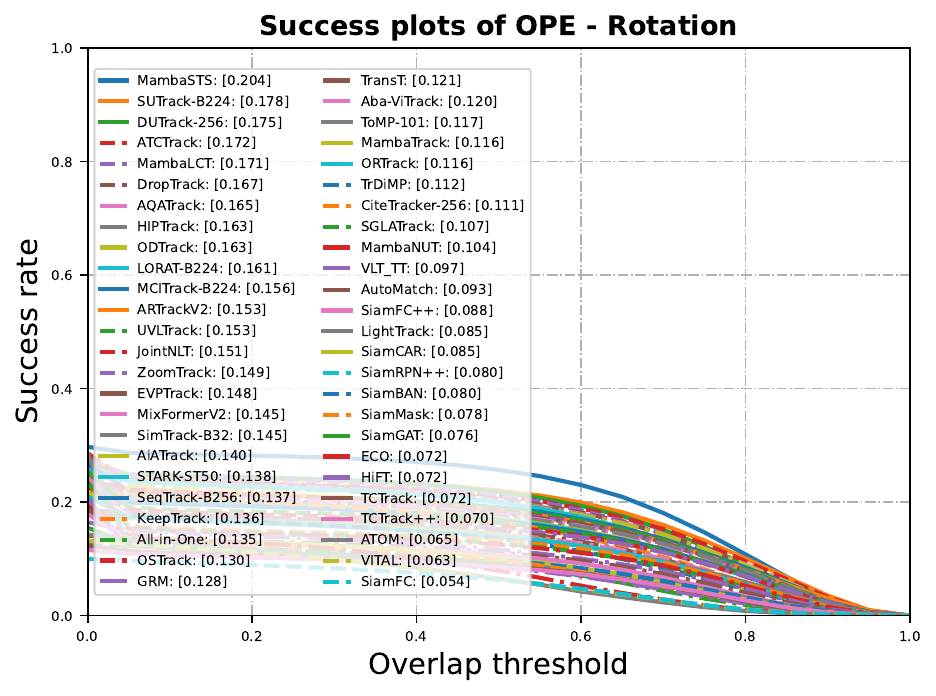}} \\

~\subfloat{\includegraphics[width =0.25\linewidth]{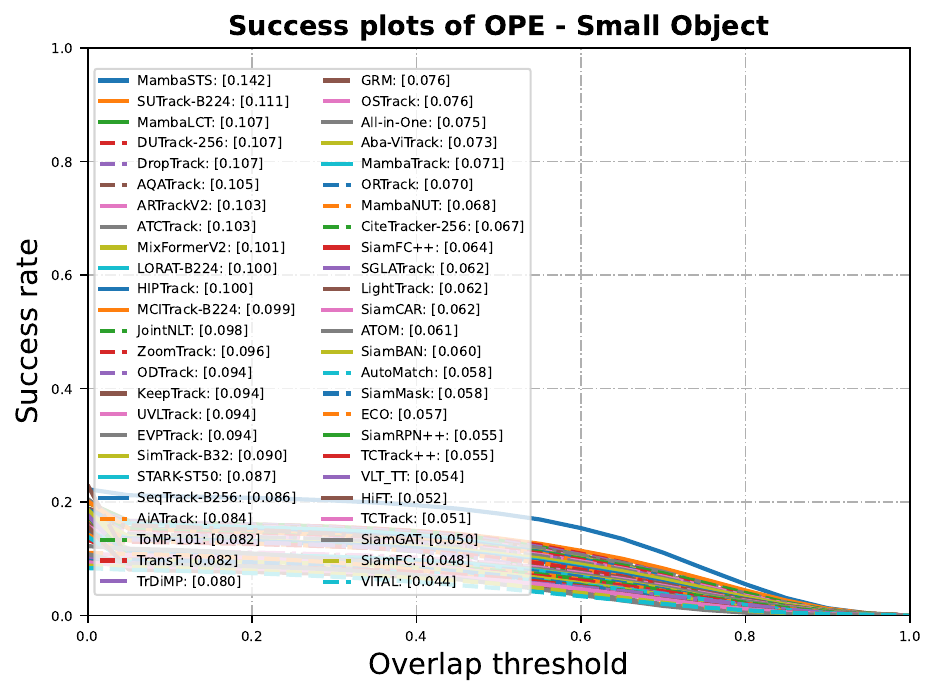}} 
~\subfloat{\includegraphics[width =0.25\linewidth]{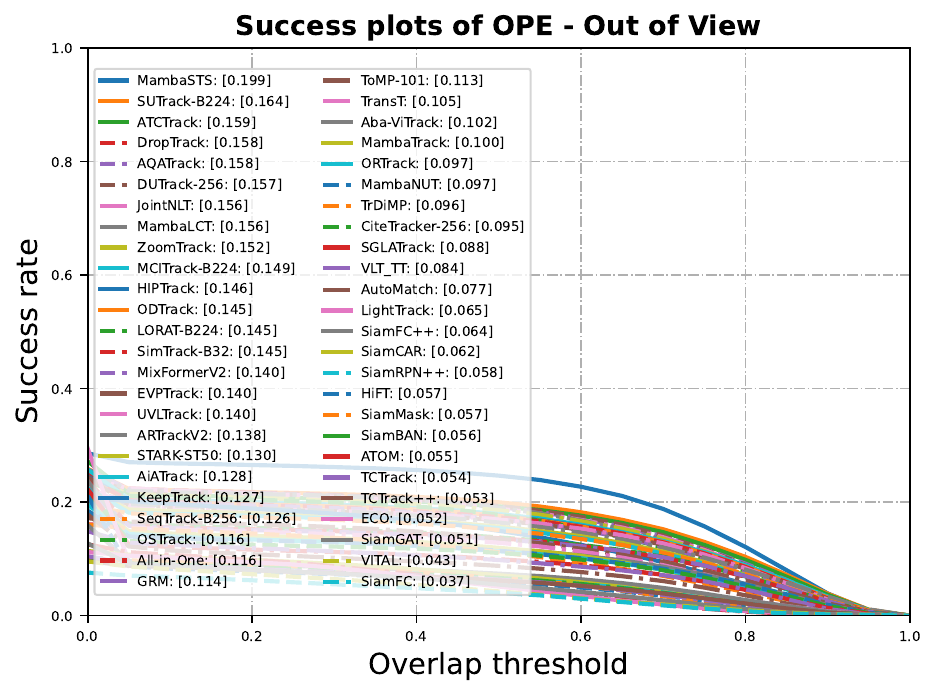}}
~\subfloat{\includegraphics[width =0.25\linewidth]{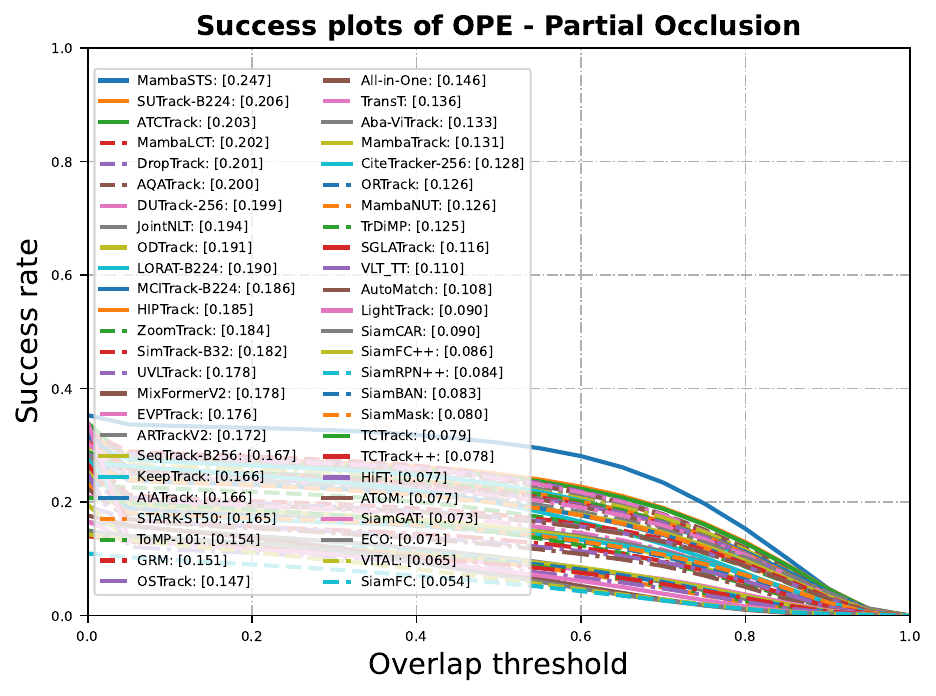}} 
~\subfloat{\includegraphics[width =0.25\linewidth]{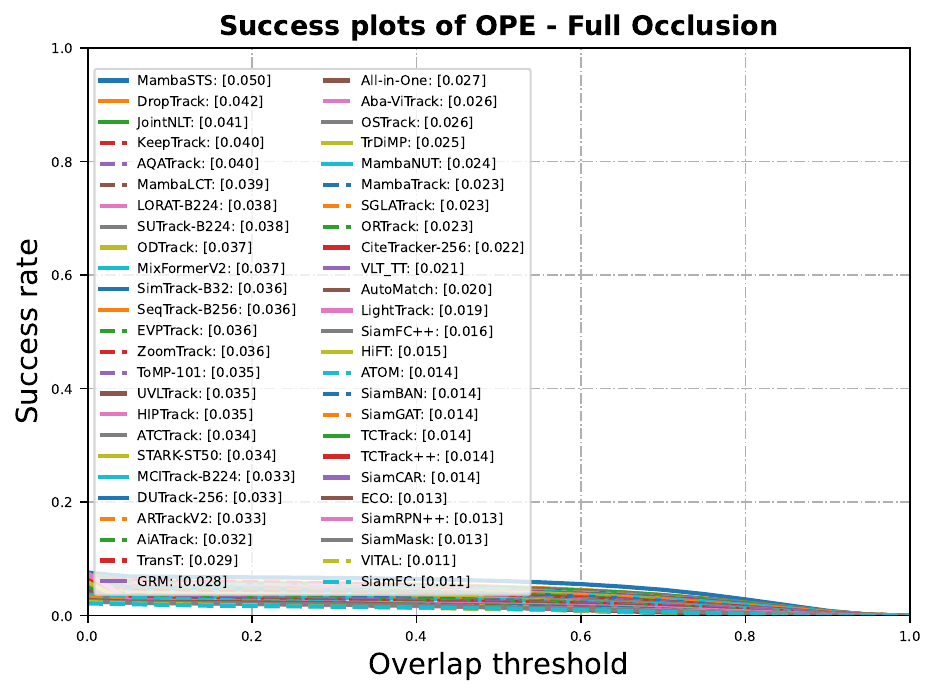}} \\

~\subfloat{\includegraphics[width =0.25\linewidth]{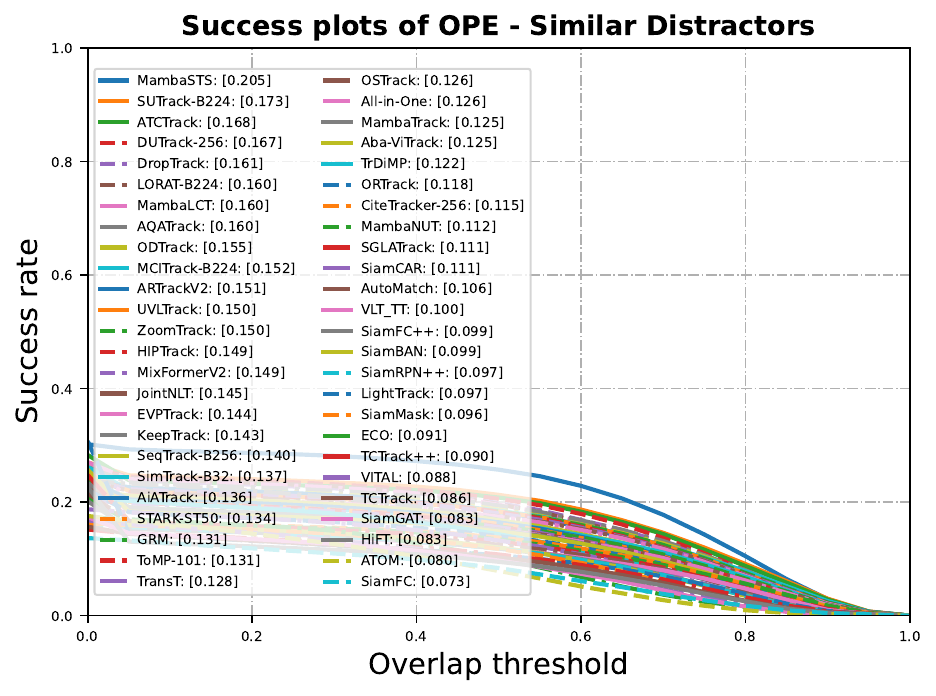}} 
~\subfloat{\includegraphics[width =0.25\linewidth]{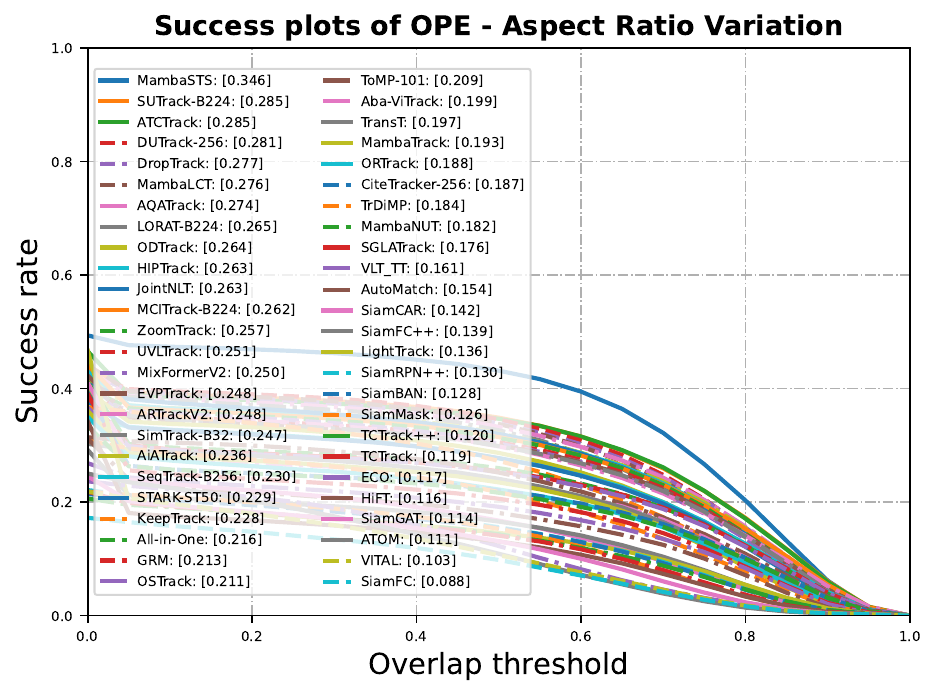}}
~\subfloat{\includegraphics[width =0.25\linewidth]{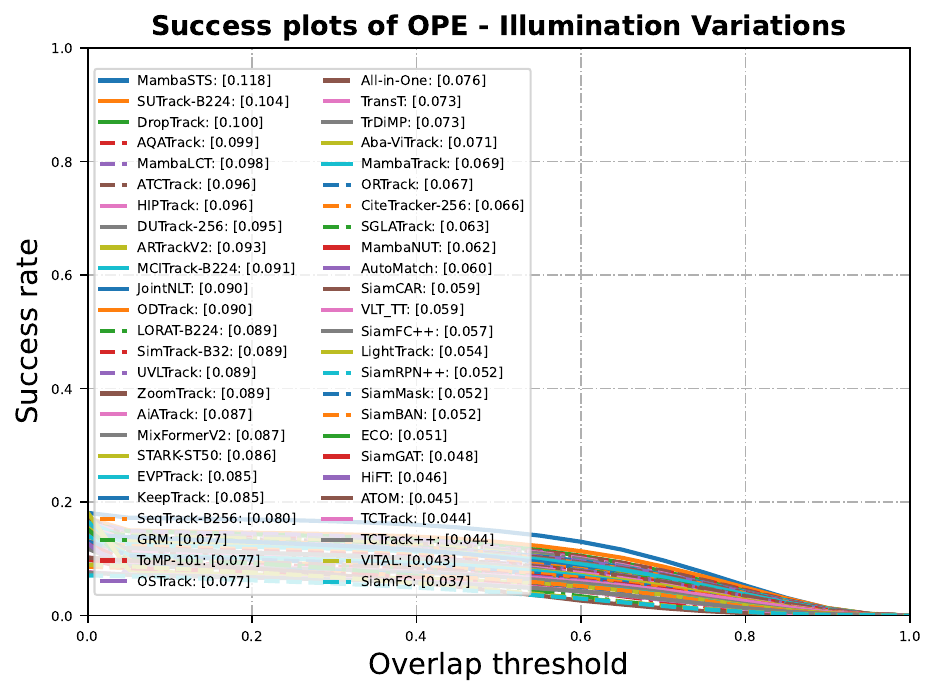}}
~\subfloat{\includegraphics[width =0.25\linewidth]{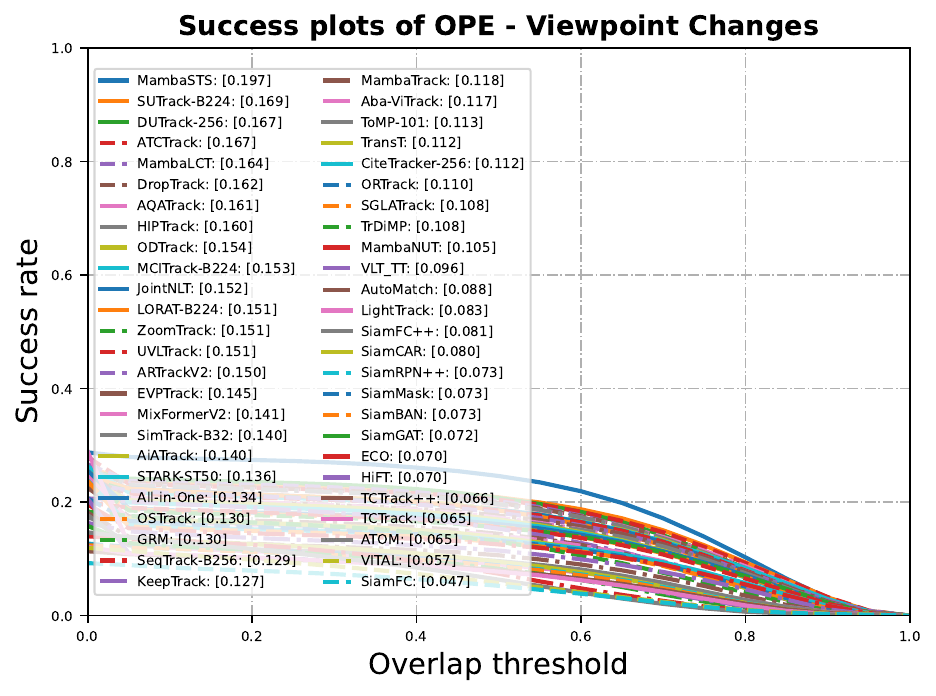}}  \\

~\subfloat{\includegraphics[width =0.25\linewidth]{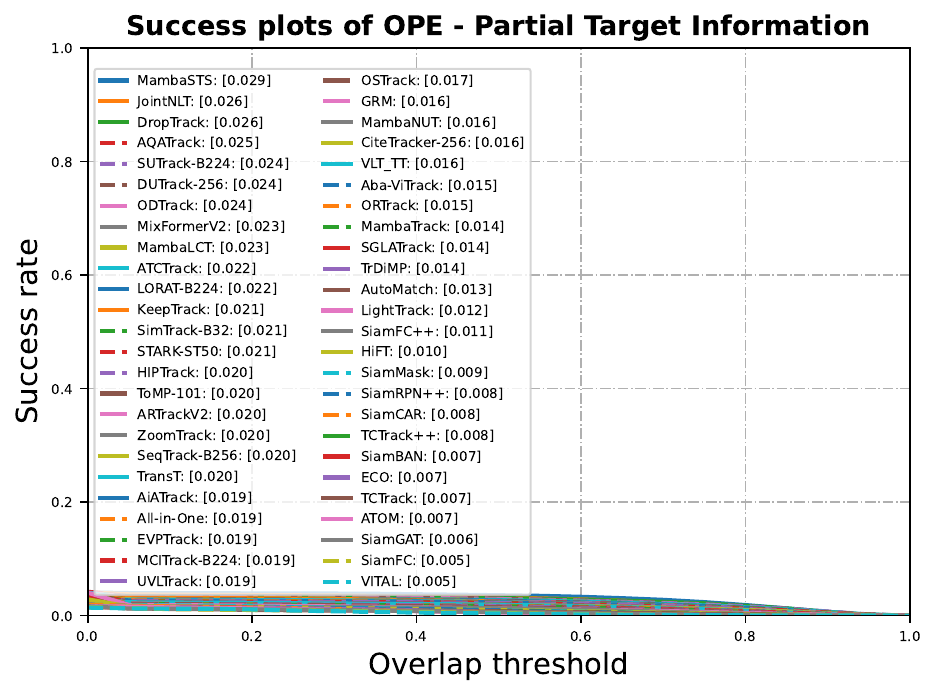}}  
~\subfloat{\includegraphics[width =0.25\linewidth]{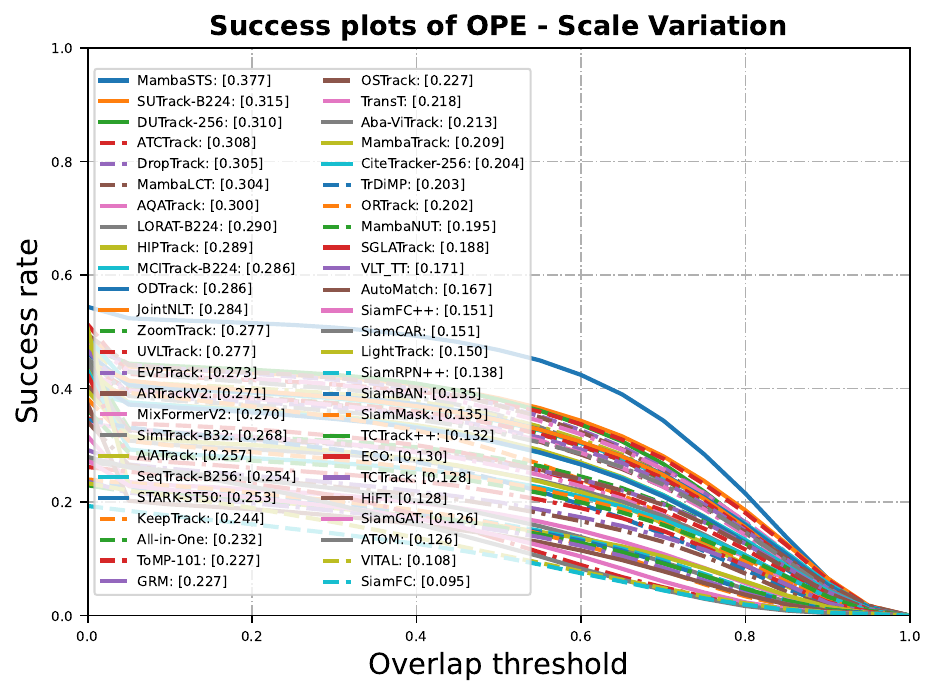}}
~\subfloat{\includegraphics[width =0.25\linewidth]{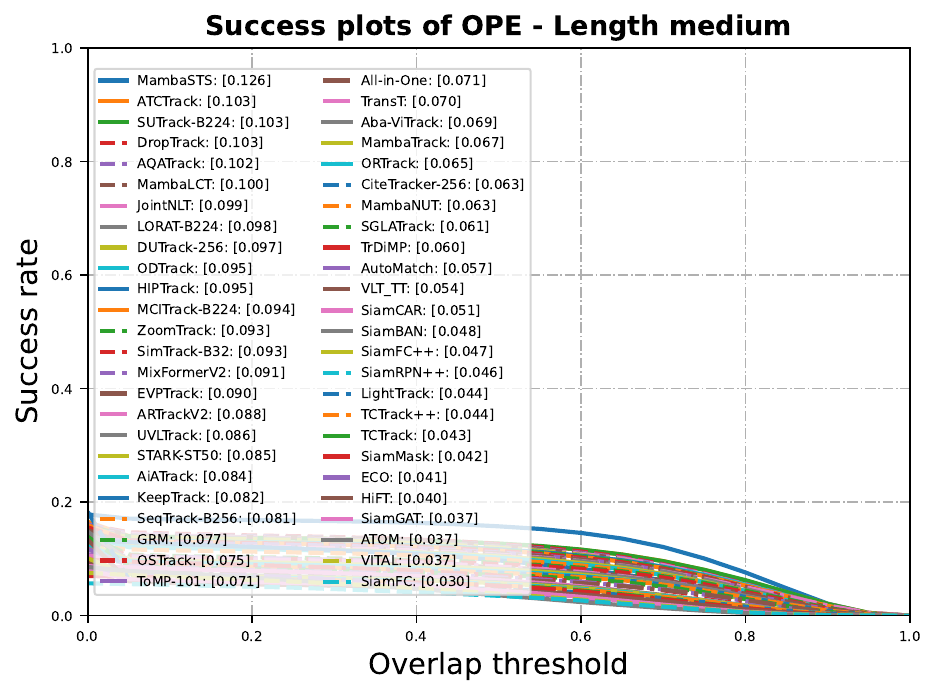}}
~\subfloat{\includegraphics[width =0.25\linewidth]{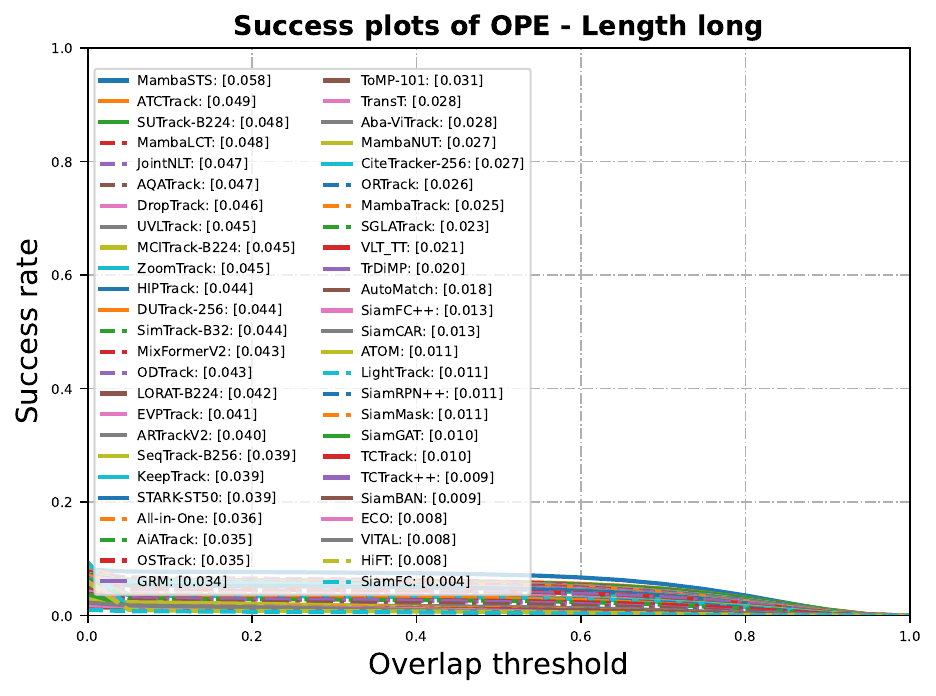}} 

\caption{Performance of 50 representative deep trackers on different tracking attributes on the UAV-Anti-UAV dataset. In each sub-figure, trackers are ranked by the success rate. Best viewed in color with zooming in.}
\label{fig:attributes_results}
\end{figure*}

\subsection{Evaluated Trackers}

To provide baseline results for future research, we extensively evaluate 50 deep trackers across five categories (see Tab.~\ref{tab:summary_of_trackers}). 

\myPara{CNN-based (15 trackers):} SiamFC~\cite{bertinetto2016fully}, ECO~\cite{danelljan2017eco}, VITAL~\cite{song2018vital}, ATOM~\cite{danelljan2019atom}, SiamMask~\cite{wang2019fast}, SiamRPN++~\cite{li2019siamrpn++}, SiamFC++~\cite{XuWLYY20}, SiamBAN~\cite{ChenZLZJ20}, SiamCAR~\cite{guo2020siamcar}, LightTrack~\cite{yan2021lighttrack}, SiamGAT~\cite{guo2021graph}, KeepTrack~\cite{mayer2021learning}, AutoMatch~\cite{zhang2021learn}, TCTrack~\cite{cao2022tctrack}, and TCTrack++~\cite{cao2023towards}.

\myPara{CNN-Transformer-based (7 trackers):} TrDiMP~\cite{wang2021transformer}, TransT~\cite{chen2021transformer}, STARK-ST50~\cite{yan2021learning}, HiFT~\cite{cao2021hift}, ToMP-101~\cite{mayer2022transforming},  AiATrack~\cite{gao2022aiatrack}, and VLT$_{\rm TT}$~\cite{guo2022divert}.

\myPara{Transformer-based (23 trackers):} SimTrack-B32~\cite{chen2022backbone}, OSTrack~\cite{ye2022joint}, ZoomTrack~\cite{kou2023zoomtrack}, MixFormerV2-B~\cite{cui2023mixformerv2}, Aba-ViTrack~\cite{li2023adaptive}, GRM~\cite{gao2023generalized}, DropTrack~\cite{wu2023dropmae}, SeqTrack-B256~\cite{chen2023seqtrack}, ODTrack~\cite{zheng2024odtrack}, EVPTrack~\cite{shi2024explicit}, AQATrack~\cite{xie2024autoregressive}, HIPTrack~\cite{cai2024hiptrack}, ARTrackV2~\cite{bai2024artrackv2}, LORAT-B224~\cite{lin2024tracking}, SGLATrack~\cite{xue2025similarity}, ORTrack~\cite{wu2025learning}, DUTrack-256~\cite{li2025dynamic}, JointNLT~\cite{zhou2023joint}, CiteTracker-256~\cite{li2023citetracker}, All-in-One~\cite{zhang2023all},  UVLTrack~\cite{ma2024unifying}, SUTrack-B224~\cite{chen2025sutrack}, and ATCTrack~\cite{feng2025atctrack}.

\myPara{Mamba-based (2 trackers):} MambaNUT~\cite{wu2024mambanut} and MambaTrack~\cite{zhang2025mambatrack}.

\myPara{Mamba-Transformer-based (3 trackers):} MambaLCT~\cite{li2025mambalct}, MCITrack-B224~\cite{kang2025exploring}, and MambaSTS (Ours).

Among these trackers, there are 10 VL tracking algorithms (\ie, VLT$_{\rm TT}$~\cite{guo2022divert}, JointNLT~\cite{zhou2023joint}, CiteTracker-256~\cite{li2023citetracker}, All-in-One~\cite{zhang2023all}, UVLTrack~\cite{ma2024unifying}, MambaTrack~\cite{zhang2025mambatrack}, SUTrack-B224~\cite{chen2025sutrack}, DUTrack-256~\cite{li2025dynamic}, ATCTrack~\cite{feng2025atctrack}, and MambaSTS (Ours)).

\myPara{\emph{Remark 1:}} The 50 evaluated algorithms span a decade and were developed across diverse platforms and deep learning frameworks. Existing research practice indicates that the performance of a single algorithm can vary significantly under different experimental environments. To ensure fairness, we strove to maintain the experimental settings of the original open-source codes and evaluated all algorithms on the same experimental platform. Furthermore, all trackers use default hyperparameters from their official implementations to ensure fair comparison.

\subsection{Evaluation Results}

\myPara{Overall Performance.} Fig.~\ref{fig:overall_performance} illustrates the performance of the 50 evaluated trackers across four key metrics—AUC, Pre, cAUC, and nPre—while Fig.~\ref{fig:mACC} presents their mACC scores, leading to several key findings: (1) MambaSTS achieves SOTA performance by outperforming all other trackers across every metric, with an AUC of 0.437, Pre of 0.602, cAUC of 0.433, nPre of 0.480, and an mACC of 0.443—this mACC score is 6.6 percentage points higher than the second-ranked SUTrack-B224 (0.377) and nearly 17 percentage points above the average mACC of 0.272 across all 50 trackers, fully validating the effectiveness of integrating spatial-temporal-semantic learning for addressing the unique challenges of UAV-Anti-UAV tracking. (2) VL-based and Mamba-based trackers stand out as the top performers alongside MambaSTS, including SUTrack-B224 (AUC=0.373, mACC=0.377), DUTrack-256 (AUC=0.364, mACC=0.369), and MambaLCT (AUC=0.357, mACC=0.360), indicating that leveraging semantic information and long-term temporal modeling is critical for handling dynamic aerial scenarios and achieving consistent overlap accuracy across frames. (3) In contrast, traditional CNN-based trackers such as SiamFC, HiFT, and TCTrack lag significantly behind with AUC values below 0.27 and mACC scores ranging from 0.135 to 0.173, primarily due to their inherent inability to model global context or dual-dynamic motion, resulting in unstable overlap performance across the highly dynamic sequences. (4) Additionally, a significant performance gap is evident: the average AUC of the 50 trackers is only around 0.30, and the average mACC is merely 0.272, both much lower than the performance observed on conventional tracking benchmarks~\cite{fan2019lasot,huang2019got}, underscoring the substantial difficulty of the UAV-Anti-UAV task and the pressing need for more specialized tracking methods.

\myPara{Attribute-based Performance.} To comprehensively analyze the trackers facing different tracking attributes, we conduct an attribute-based evaluation using 15 tracking attributes. The results are shown in Fig.~\ref{fig:attributes_results}. Key observations: (1) MambaSTS is robust to diverse challenges: It maintains the highest success rate across critical attributes, including fast motion (FM), motion blur (MB), small object (SO), and similar distractors (SD). For example, MambaSTS achieves a success rate of 0.205 on SD, compared to 0.125 for MambaTrack and 0.118 for ORTrack. (2) Existing trackers struggle with illumination and occlusion: For attributes like illumination variations (IV) and full occlusion (FO), even top trackers (including MambaSTS, SUTrack-B224, and ATCTrack) achieve success rates $<$0.15, indicating a lack of effective adaptation mechanisms. (3) Motion-related attributes are bottlenecks: Attributes like CM, FM, and ROT are very challenging, with average success rates $<$0.3, emphasizing the need for improved motion modeling.

\begin{figure*}[ht]
\centering
\subfloat{\includegraphics[width =0.5\columnwidth]{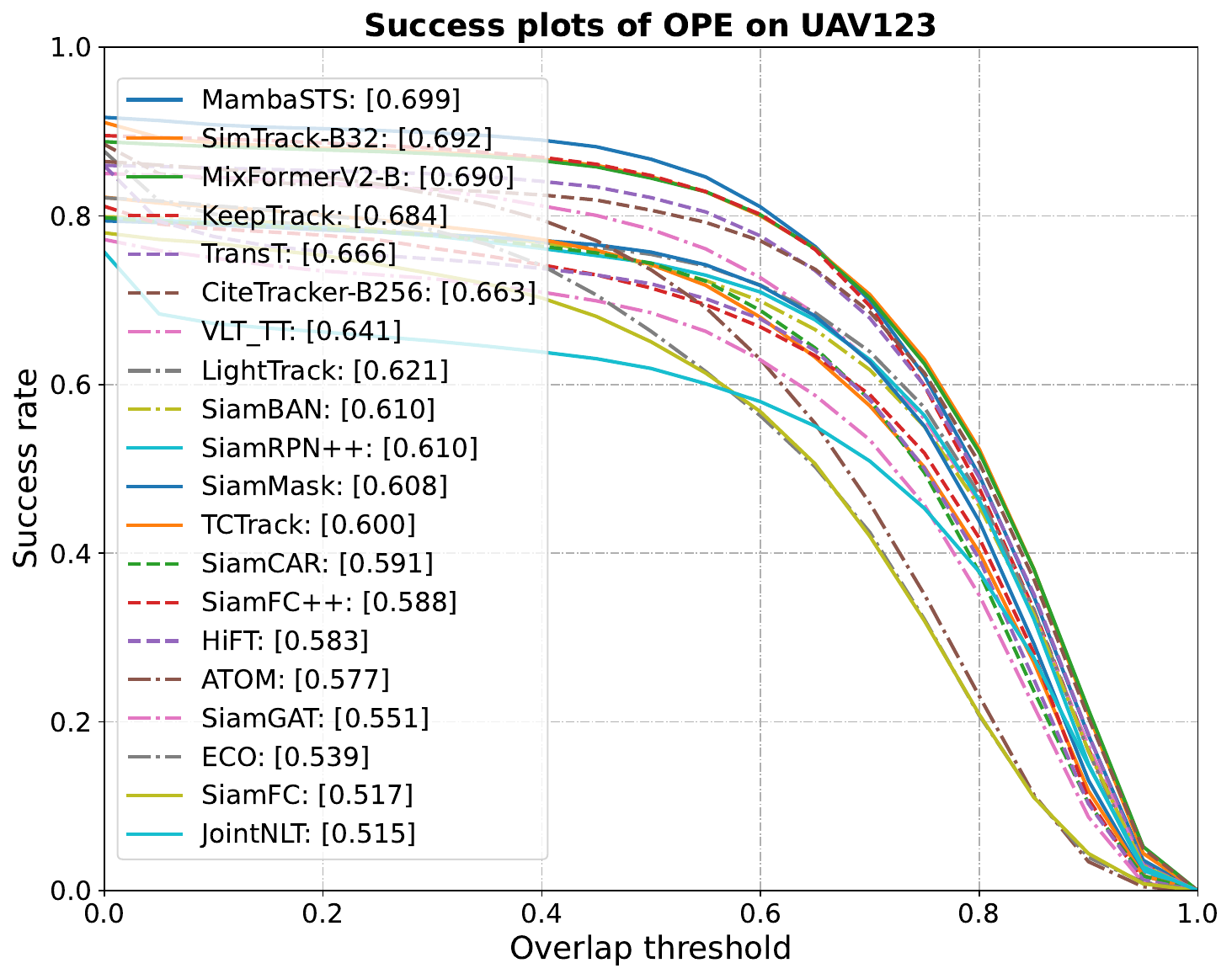}}~
\subfloat{\includegraphics[width =0.5\columnwidth]{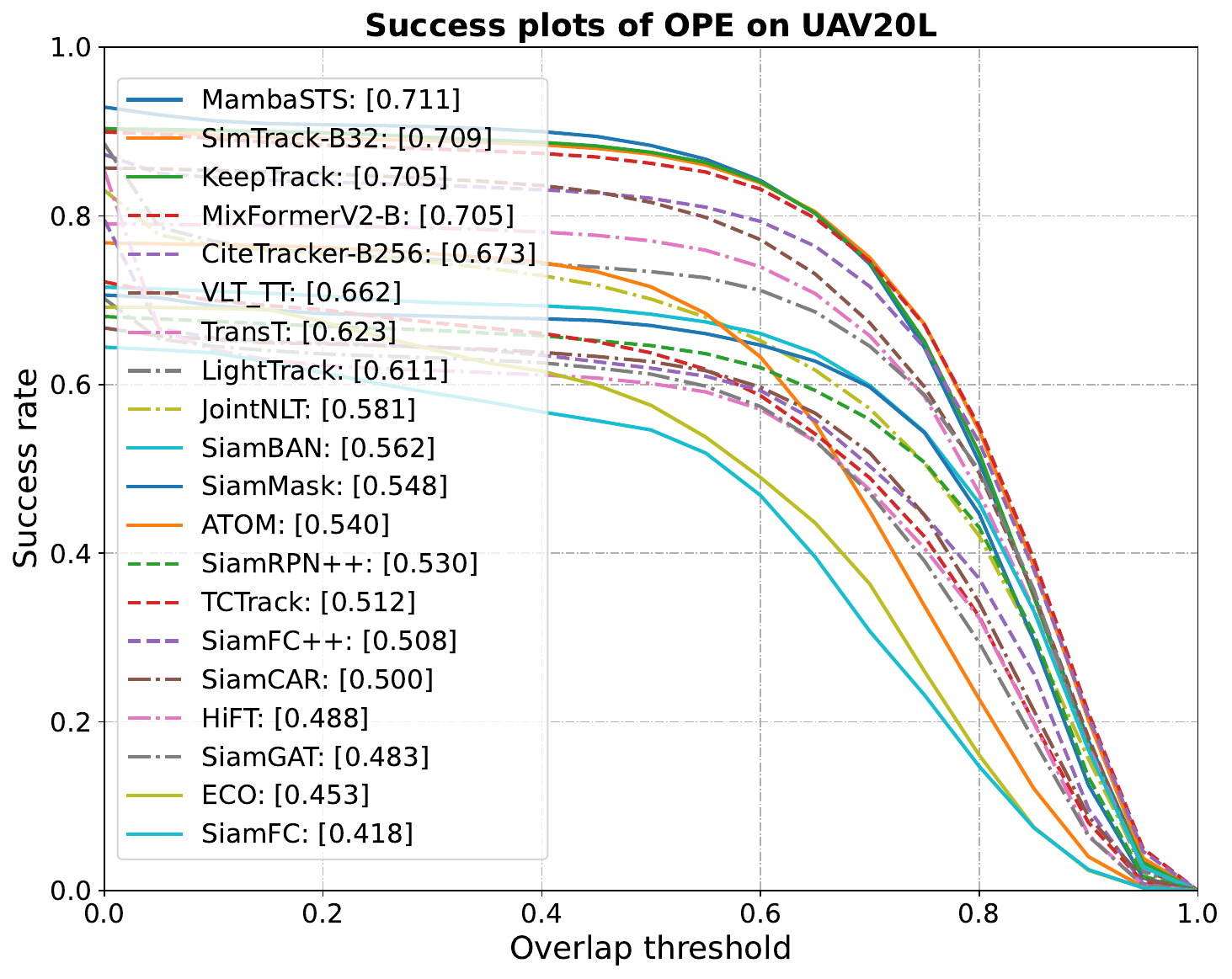}}~
\subfloat{\includegraphics[width =0.5\columnwidth]{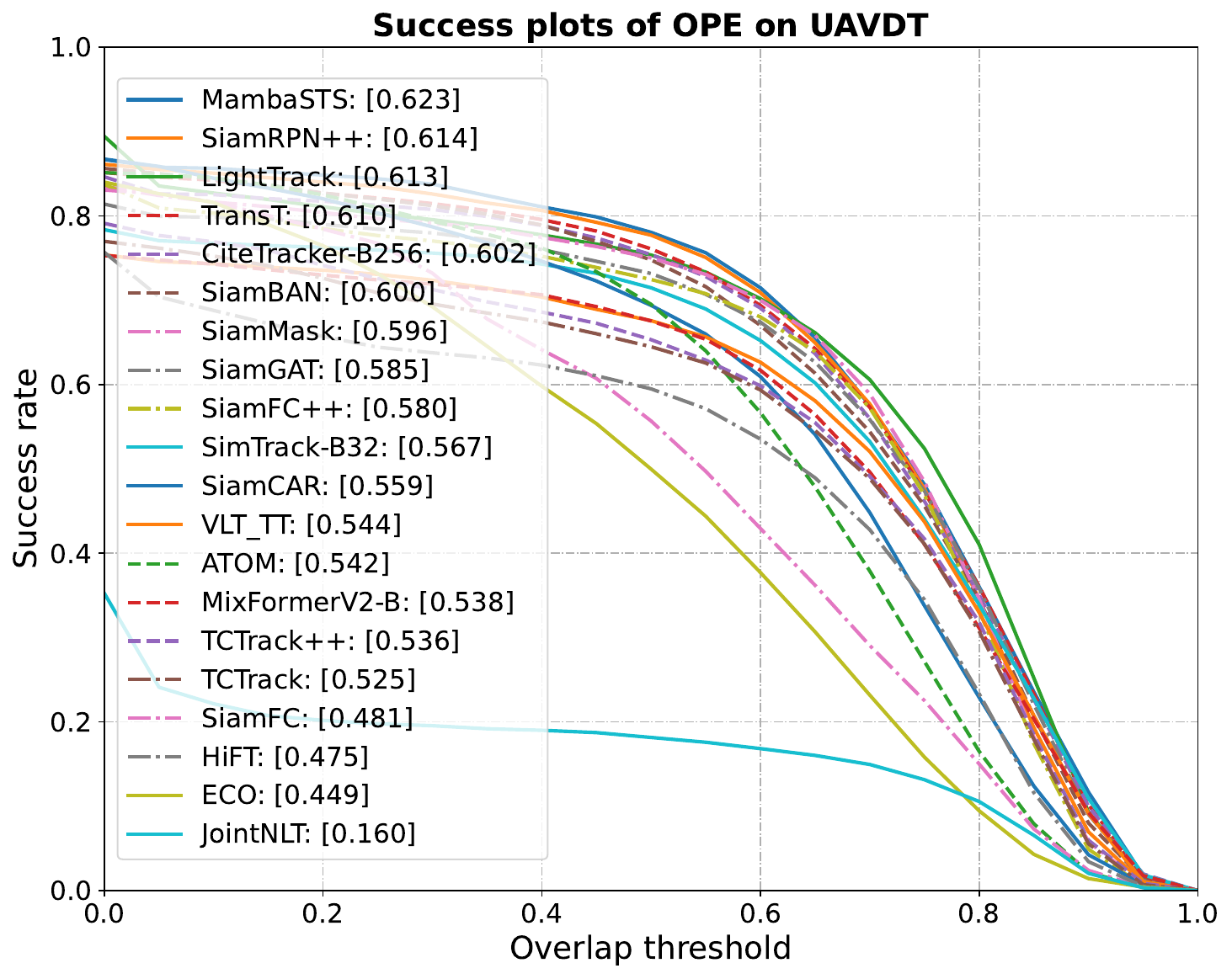}}~
\subfloat{\includegraphics[width =0.5\columnwidth]{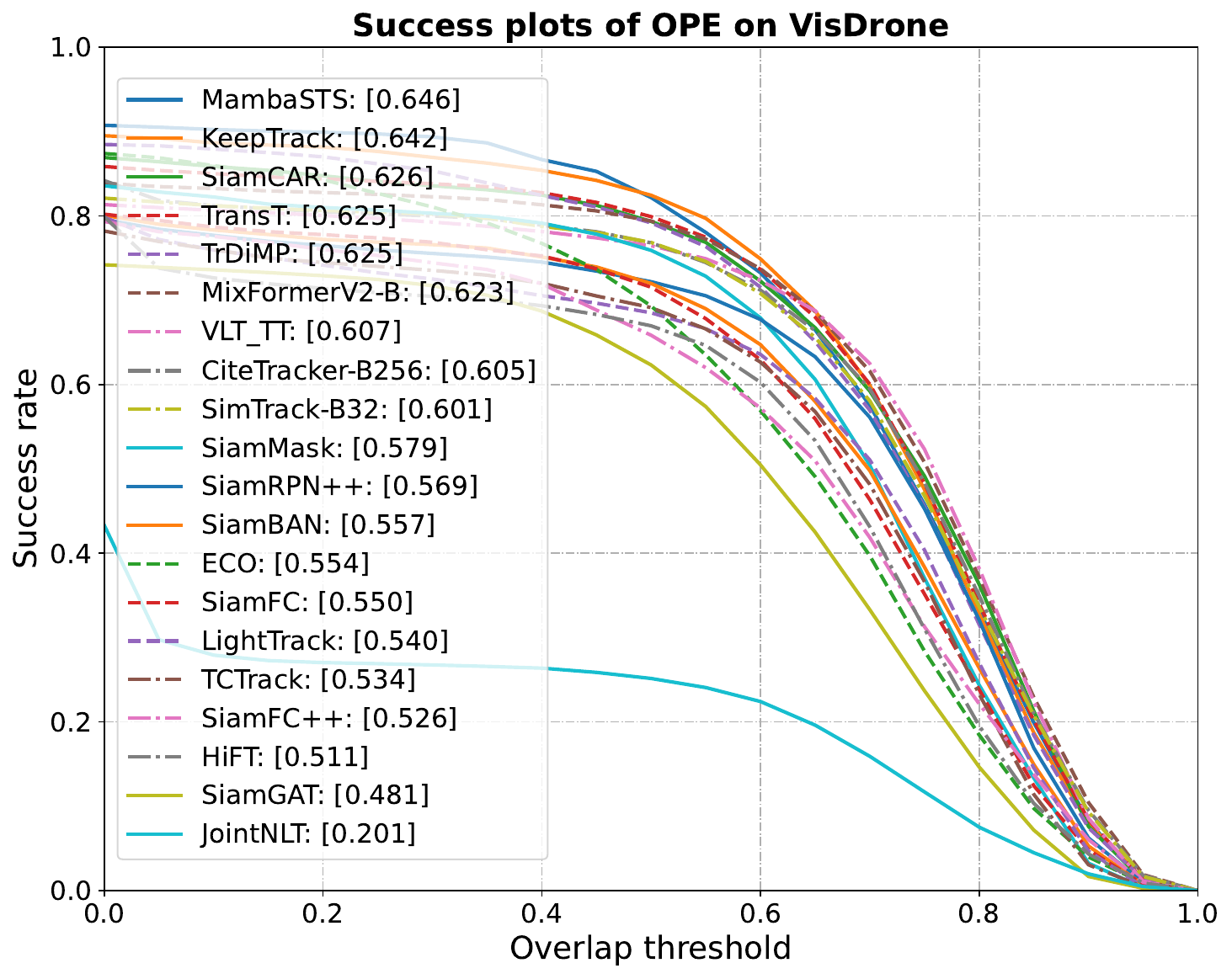}}\\
\caption{Comparison with SOTA trackers on four UAV tracking datasets (\ie, UAV123, UAV20L, UAVDT, and VisDrone).}
\label{fig:four_uav_tracking_results}
\end{figure*}

\begin{figure}[ht]
\centering
\subfloat{\includegraphics[width =0.5\columnwidth]{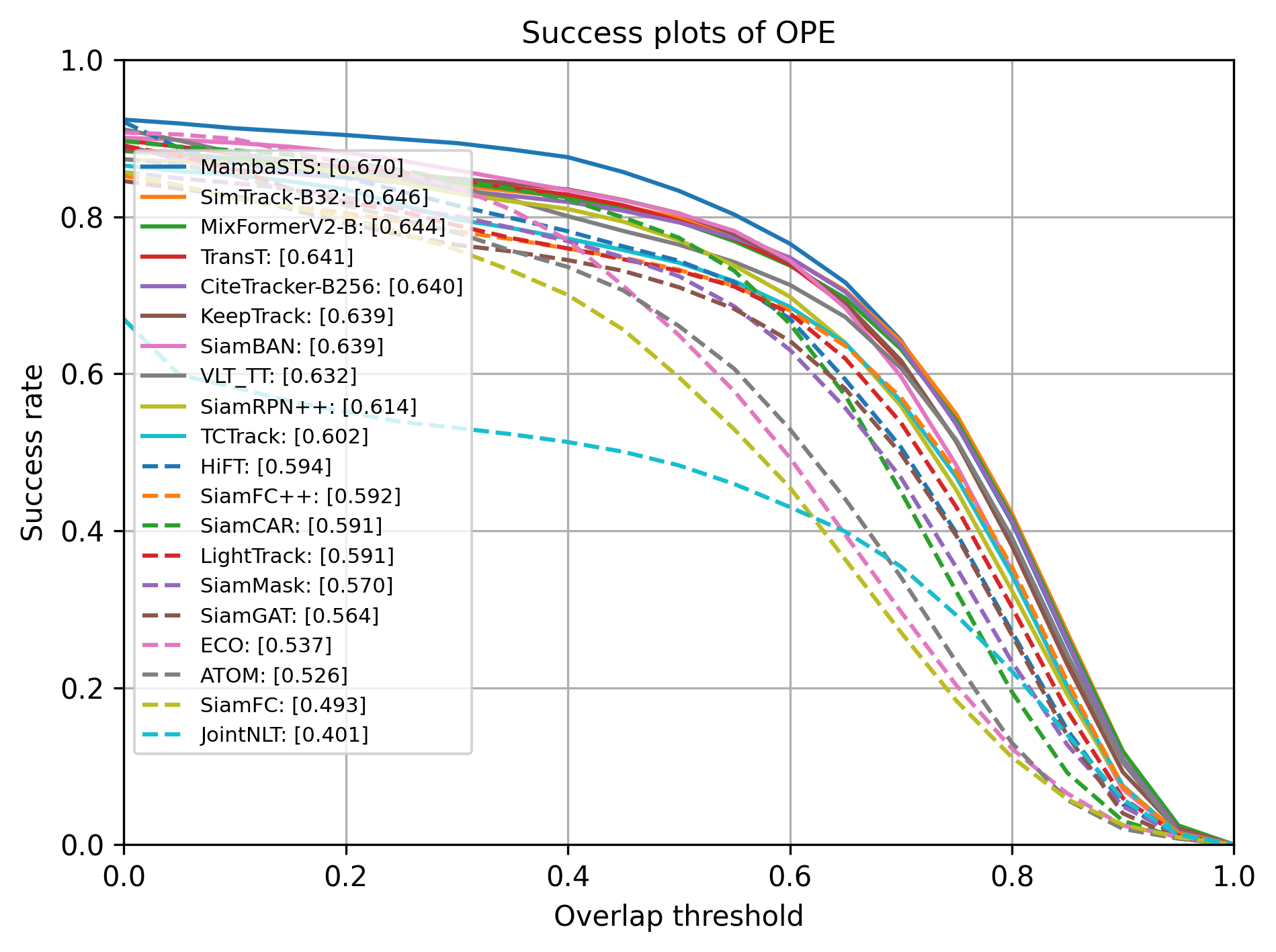}}~
\subfloat{\includegraphics[width =0.5\columnwidth]{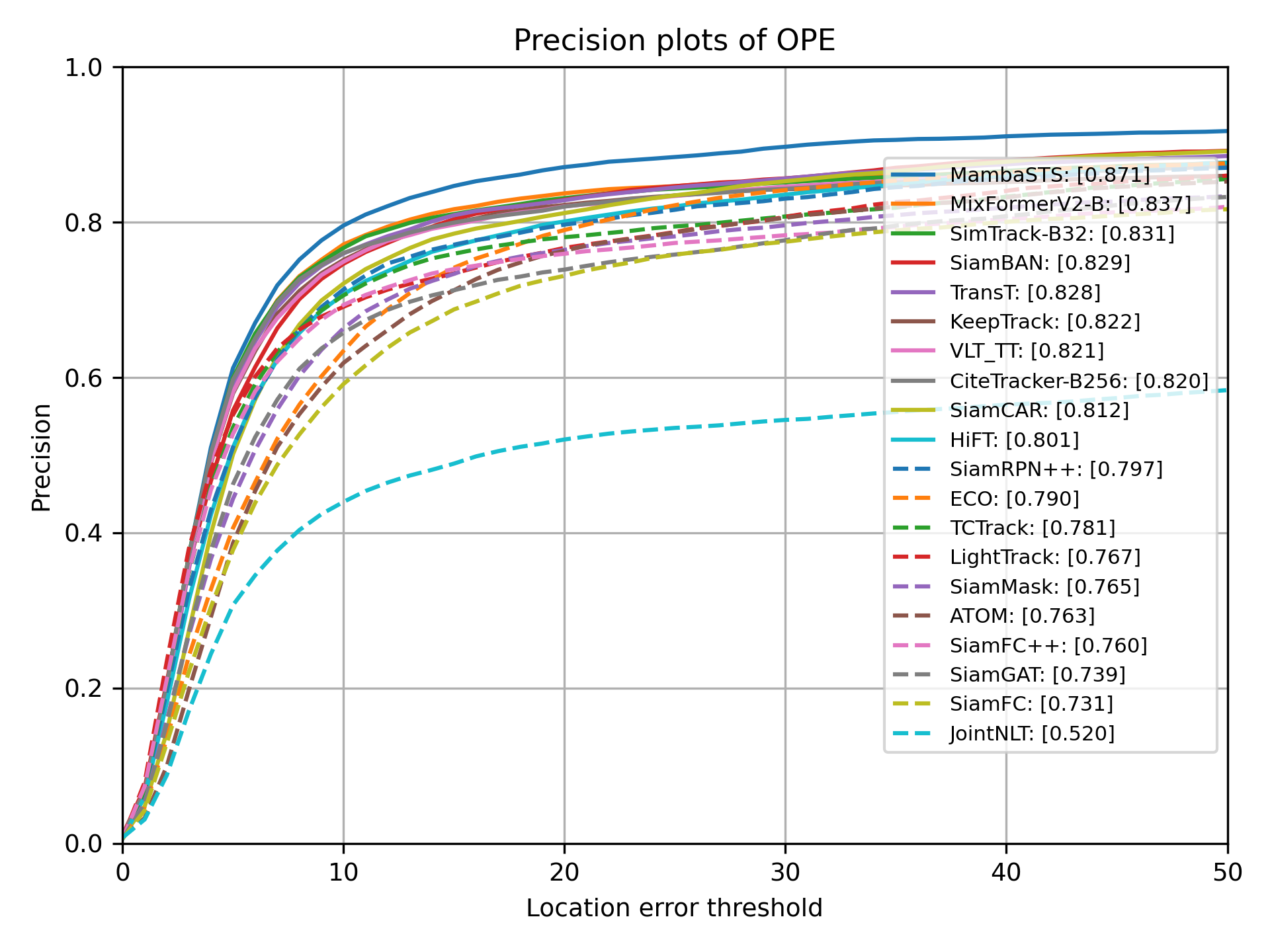}}\\
\caption{Comparison with SOTA trackers on the DBT70 dataset.}
\label{fig:dtb70}
\end{figure}

\subsection{Generalization to Existing Benchmarks}
To further validate the generalization capability of MambaSTS beyond the proposed UAV-Anti-UAV task, we evaluate it on two representative tracking tasks (UAV tracking and ground-based Anti-UAV tracking).

\myPara{Results on UAV Tracking Datasets.} To verify MambaSTS’s generalization to UAV tracking tasks focusing on ground targets, we conduct evaluations on five representative datasets: UAV123, UAV20L, UAVDT, VisDrone, and DBT70 (Figs.~\ref{fig:four_uav_tracking_results} and~\ref{fig:dtb70}). These datasets cover diverse scenarios such as urban areas, rural landscapes, and varying illumination conditions, presenting typical challenges of ground-target UAV tracking, including top-down viewpoints, scale variations, and motion blur. The results demonstrate that MambaSTS achieves SOTA performance across all five datasets, with an AUC of 0.699 on UAV123, 0.711 on UAV20L, 0.623 on UAVDT, 0.646 on VisDrone, and competitive performance on DBT70. It outperforms leading trackers such as SimTrack-B32, KeepTrack, and MixFormerV2-B, confirming that the spatial-temporal-semantic learning mechanism in MambaSTS is not only tailored for air-to-air dynamic tracking but also effectively adapts to the unique characteristics of UAV-mounted camera capture for ground targets.

\myPara{Results on Anti-UAV Tracking Datasets.} We also assess MambaSTS’s generalization to ground-based Anti-UAV tracking tasks by testing it on two widely used datasets: Anti-UAV318 and DUT Anti-UAV (see Tab.~\ref{tab:antiuav_tracking_results}). These datasets consist of RGB and infrared video sequences captured by fixed ground cameras, focusing on core challenges like small targets, cluttered sky backgrounds, and low illumination that are inherent to ground-based Anti-UAV scenarios. MambaSTS achieves the highest performance on both datasets, delivering an AUC of 62.5\% and Pre of 84.4\% on Anti-UAV318, as well as an AUC of 61.3\% and Pre of 82.2\% on DUT Anti-UAV. Compared to the previous SOTA method MambaTrack, MambaSTS improves the AUC by 0.6\% and Pre by 6.1\% on Anti-UAV318, and the AUC by 3.2\% and Pre by 1.9\% on DUT Anti-UAV. This validates MambaSTS’s versatility in handling both air-to-air and ground-based Anti-UAV tracking tasks, solidifying its potential as a universal baseline for various low-altitude security applications.

\begin{table}[ht]
\centering
  \caption{Comparison with SOTA trackers on two Anti-UAV tracking datasets (\ie, Anti-UAV318 and DUT Anti-UAV).}
  \label{tab:antiuav_tracking_results}
  \centering
  \setlength{\tabcolsep}{2.8mm}{
  \scalebox{1.0}{
  \begin{tabular}{lcccc}
    \Xhline{0.75pt} 
      \multicolumn{1}{l}{\multirow{2}[1]{*}{Method}} &  \multicolumn{2}{c}{ Anti-UAV318}  & \multicolumn{2}{c}{DUT Anti-UAV} \\
   
   \cmidrule(r){2-3} \cmidrule(r){4-5}  &  AUC (\%)  &  Pre (\%)    & AUC (\%)  &  Pre (\%)  \\
    \hline

    SiamMask~\cite{wang2019fast}  & 38.6 &   49.9  &   44.7  & 64.0    \\

    SiamBAN~\cite{ChenZLZJ20}  & 40.5 &   52.7  &   51.4  & 69.3    \\

    HiFT~\cite{cao2021hift}  &   41.7  &   54.2  & 39.7   & 55.3 \\

     TCTrack~\cite{cao2022tctrack}  & 42.4 &   55.2  &   46.1  & 64.1    \\

     SiamGAT~\cite{guo2021graph}  & 42.9 &   58.3  &   52.8  & 72.9    \\

      SiamFC++~\cite{XuWLYY20}  & 44.0 &   56.1  &   52.5  & 72.4    \\
      LightTrack~\cite{yan2021lighttrack}   & 44.9 &   55.9  &   47.5  & 60.6    \\
   
     SiamRPN++~\cite{li2019siamrpn++}  & 45.7 &   57.6  &   58.2  & 77.0    \\
  
   AutoMatch~\cite{zhang2021learn}  & 47.5 &   59.7  &   57.8  & 74.1    \\

    SiamCAR~\cite{guo2020siamcar}  & 48.1 &   61.4  &   52.6  & 70.3    \\
     
     Aba-ViTrack~\cite{li2023adaptive} & 53.5 &   66.5  &   60.1  & 80.0    \\
     
    MambaNUT~\cite{wu2024mambanut} & 53.9 &   66.8  &   54.0  & 69.9    \\
         
     CiteTracker-256~\cite{li2023citetracker}  & 54.5 &   67.0  &   55.3  & 71.9    \\
     SGLATrack~\cite{xue2025similarity}  & 56.7 &   71.2  &   56.0  & 75.9    \\

     ORTrack~\cite{wu2025learning}  & 58.8 &   74.1  &   57.6  & 77.6    \\
     MambaTrack~\cite{zhang2025mambatrack}  & 61.9 &   78.3  &   58.1  & 80.3    \\

    \hline
    
     \textbf{MambaSTS (Ours)}  & \textbf{62.5} &   \textbf{84.4}  &   \textbf{61.3}  & \textbf{82.2}    \\
    
    \Xhline{0.75pt} 

  \end{tabular}
  }}~~
\end{table}

\subsection{Ablation Study}
\label{sec:ablation}

To analyze the contribution of each component in MambaSTS, we conduct ablation experiments on the UAV-Anti-UAV dataset (see Tab.~\ref{tab:ablation}). We adopt OSTrack with a ViT-B backbone as the baseline tracker. The study progressively incorporates the three core components: (1) \textbf{Temporal Modeling}, via the long-term context module and STS Mamba module; (2) \textbf{Spatial Modeling}, by replacing the standard ViT backbone with the hierarchical HiViT encoder; and (3) \textbf{Semantic Modeling}, by integrating the language encoder and semantic guidance.

\myPara{Effectiveness of Temporal Modeling.}
As shown in the first two rows of Tab.~\ref{tab:ablation}, the baseline tracker achieves an AUC score of 27.8\% and a Pre score of 38.3\%. By introducing Temporal Modeling, specifically the temporal token propagation mechanism that captures video-level context, the performance improves significantly. The AUC increases by 5.6\% (from 27.8\% to 33.4\%), and Pre improves by 8.4\% (from 38.3\% to 46.7\%). This demonstrates that modeling long-term dependencies is crucial for handling the dual-dynamic disturbances in aerial tracking, allowing the tracker to maintain a memory of the target state beyond the current frame.

\myPara{Effectiveness of Spatial Modeling.}
The third row validates the impact of our Spatial Modeling strategy. By replacing the standard ViT backbone with the HiViT encoder (designed for global spatial aggregation), the tracker gains further improvements, boosting AUC to 36.9\% and Pre to 53.1\%. This represents a steady gain of 3.5\% in AUC over the temporal-only version. The hierarchical nature of HiViT allows the model to better handle the extreme scale variations and small targets characteristic of the UAV-Anti-UAV dataset, providing more robust visual feature representations.

\myPara{Effectiveness of Semantic Modeling.}
Finally, the integration of Semantic Modeling (Row 4) completes the MambaSTS framework. Adding the language prompt and semantic fusion yields substantial performance gains, increasing the AUC to 43.7\% and Pre to 60.2\%. Compared to the version without semantics, this provides a massive 6.8\% improvement in AUC. This result confirms that natural language descriptions serve as a strong prior, enabling the tracker to distinguish the target from similar distractors and background clutter where visual features alone might be ambiguous.

In summary, each component plays an indispensable role. The complete MambaSTS framework achieves a total improvement of 15.9\% in AUC compared to the baseline (27.8\% vs. 43.7\%), validating the effectiveness of our unified spatial-temporal-semantic learning approach.

\subsection{Computational Complexity}

We evaluate MambaSTS’s runtime on a single NVIDIA A6000 GPU. It achieves 54 FPS, which is comparable to real-time trackers (\eg, OSTrack=105 FPS, and MixFormerV2-B=165 FPS) and meets the requirements of UAV-Anti-UAV applications ($\geq$30 FPS). The balance between performance and efficiency makes MambaSTS a practical baseline for aerial tracking.

\subsection{Limitations and Discussion}

This work provides a foundational benchmark and baseline for UAV-Anti-UAV tracking, yet several limitations point to future research directions.

First, while our dataset provides a significant step forward in scale and realism, it primarily consists of visible-light (RGB) videos. In real-world security operations, especially in nighttime or adverse weather conditions, multi-modal sensing (\eg, infrared, LiDAR) is often critical. Future work could expand the benchmark to include such modalities. Furthermore, although we collect various flight dynamics, the dataset cannot fully encompass the infinite variability of real-world adversarial flight maneuvers and physical interactions (\eg, collisions, electronic countermeasures), which remain challenging to capture at scale.

Second, MambaSTS, while effective, inherits certain constraints. Its performance, though superior, still leaves considerable room for improvement—the overall AUC of 43.7\% underscores the intrinsic difficulty of the UAV-Anti-UAV task. The model’s dependency on a textual prompt, while beneficial for semantics, introduces a requirement for accurate language description, which may not always be available in fully autonomous interception systems. Additionally, the current architecture is trained offline; extending it to an online adaptation framework that continuously learns from new observations during pursuit could further enhance robustness in dynamic, unseen scenes.

Despite these limitations, the proposed UAV-Anti-UAV benchmark and MambaSTS baseline provide a foundational framework for future research, shedding light on the unique challenges of air-to-air tracking and offering actionable insights for developing next-generation anti-UAV systems. As low-altitude security threats continue to evolve, addressing these limitations will be crucial for translating academic research into practical, reliable, and robust anti-UAV technologies.

\begin{table}[t]
  \centering
  \caption{Ablation study on the UAV-Anti-UAV dataset. The baseline is OSTrack with a ViT-B backbone. We progressively add Temporal, Spatial, and Semantic modeling components to validate their effectiveness.}
  \label{tab:ablation}
  \setlength{\tabcolsep}{1.4mm}{
  \scalebox{1.0}{
  \begin{tabular}{ccc|cccccccc}
    \Xhline{0.75pt} 
     Temporal  & Spatial & Semantic  & AUC (\%) & Pre (\%)   & nPre (\%)  & cAUC (\%)  \\
    \hline

    - &   - &  -   &   27.8   &   38.3  &  30.6  &   27.1  \\

    \cmark & -  & - &  33.4   &  46.7  & 37.4  & 32.8    \\

    \cmark  &   \cmark  & -  &  36.9   &  53.1  & 41.8  & 36.5   \\

    \cmark  &   \cmark  &   \cmark &  \textbf{43.7}   &  \textbf{60.2}  &  \textbf{48.0}  &  \textbf{43.3}   \\
    
    \Xhline{0.75pt} 
  \end{tabular}
  }}
\end{table} 

\section{Conclusion}
\label{sec:conclusion}

This paper takes a pioneering step towards practical aerial safety by introducing the UAV-Anti-UAV tracking task. Distinct from existing benchmarks limited to ground-based sensors, our work addresses the complexities of air-to-air tracking, characterized by the rapid motion of both the shooting platform and the target. To the best of our knowledge, we present the first million-scale multi-modal benchmark for this domain, comprising 1.05 million frames across 1,810 videos, which we believe will serve as a standard testbed for the community. Alongside the dataset, we propose MambaSTS, a novel tracking baseline that integrates spatial, temporal, and semantic learning through a state-space modeling mechanism, effectively capturing long-term dependencies in aerial video sequences. Comprehensive evaluations of 50 modern trackers on our benchmark reveal significant performance gaps, underscoring the difficulty of this task and the limitations of current methods in handling dynamic aerial interactions. We hope these contributions will significantly accelerate the development of next-generation anti-UAV systems and inspire further research in multi-modal aerial tracking.

\bibliographystyle{IEEEtran}
\bibliography{main}

\end{document}